\PassOptionsToPackage{shortlabels}{enumitem}
\documentclass[11pt, a4paper]{sakana_format}
\usepackage{xurl}
\usepackage[utf8]{inputenc}
\usepackage{graphicx}
\usepackage{booktabs}
\usepackage{multirow}
\usepackage{amsmath}
\usepackage{amssymb}
\usepackage{colortbl}
\usepackage[whole]{bxcjkjatype}
\definecolor{lstframe}{RGB}{70,130,180}
\definecolor{lstbg}{RGB}{240,247,255}
\definecolor{exframe}{RGB}{60,140,90}
\definecolor{exbg}{RGB}{241,250,244}   %
\definecolor{trajframe}{RGB}{120,90,190}
\definecolor{trajbg}{RGB}{246,244,252} %
\definecolor{deltagreen}{RGB}{0,158,115} %
\usepackage{tcolorbox}
\tcbuselibrary{breakable}
\usepackage{fontawesome5}
\tcbuselibrary{skins}
\definecolor{envframe}{RGB}{150,150,150}
\definecolor{sumbg}{RGB}{253,236,224}
\definecolor{sumframe}{RGB}{200,120,70}
\tcbset{trajbase/.style={enhanced, boxrule=0.5pt, arc=1pt, left=2.5pt, right=2.5pt, top=1.5pt, bottom=2pt,
  before skip=3pt, after skip=3pt}}
\newcommand{\trajtitle}[1]{\raggedright{\bfseries #1}\par\vspace{1pt}}
\newtcolorbox{trajagent}[1]{trajbase, colback=trajbg, colframe=trajframe, before upper={\trajtitle{\faRobot\ Agent\hfill{\normalfont\color{black!55}#1}}}}
\newtcolorbox{trajenv}{trajbase, colback=white, colframe=envframe, before upper={\trajtitle{\faTerminal\ Environment}}}
\newtcolorbox{trajinfo}[1]{trajbase, colback=white, colframe=black!70, before upper={\trajtitle{#1}}}
\newtcolorbox{trajsum}[1]{trajbase, colback=sumbg, colframe=sumframe, before upper={\trajtitle{#1}}}
\newtcolorbox{trajfinish}{trajbase, colback=green!8, colframe=green!45!black, before upper={\raggedright}}
\newtcolorbox{trajhead}{enhanced, boxrule=0.4pt, arc=0pt, left=1pt, right=1pt, top=0.5pt, bottom=0.5pt,
  before skip=3pt, after skip=2pt, colback=black!8, colframe=black!40, halign=center, fontupper=\bfseries\small}
\newtcolorbox{trajbox}[1]{enhanced, breakable, colback=trajbg, colframe=trajframe, boxrule=0.6pt, arc=0pt,
  colbacktitle=trajframe!15, coltitle=black, fonttitle=\bfseries, toptitle=1.5pt, bottomtitle=1.5pt,
  titlerule=0.6pt, titlerule style=trajframe, fontupper=\small, before upper={\raggedright},
  title={\colorbox{trajframe}{\textcolor{white}{\sffamily\bfseries\footnotesize TRAJECTORY}}\hspace{0.5em}#1}}
\newcommand{\trajstep}[1]{\par\smallskip\noindent\textbf{Step #1}\quad}
\newcommand{\trajcmd}[1]{\par\noindent{\ttfamily\footnotesize #1}}
\newtcolorbox{promptbox}[1]{enhanced, breakable, colback=lstbg, colframe=lstframe, boxrule=0.6pt, arc=0pt,
  colbacktitle=lstframe!18, coltitle=black, fonttitle=\bfseries, toptitle=1.5pt, bottomtitle=1.5pt,
  titlerule=0.6pt, titlerule style=lstframe,
  title={\colorbox{lstframe}{\textcolor{white}{\sffamily\bfseries\footnotesize PROMPT}}\hspace{0.5em}#1},
  segmentation style={dashed, black!40}}
\usepackage{tikz}
\usetikzlibrary{positioning, arrows.meta, fit, backgrounds, shapes.geometric}
\usepackage{pifont}
\newcommand{\cmark}{\ding{51}}
\newcommand{\xmark}{\ding{55}}
\newcommand{\method}{JFinBench}
\let\includegraphics\includegraphicsoriginal
\graphicspath{{assets/acl/}{assets/}}

\newcommand{\papercodelink}{%
  \par\smallskip
  \begin{center}
    \small
    \begin{tabular}{@{}l@{}}
      \raisebox{-0.15em}{\includegraphics[height=1.1em]{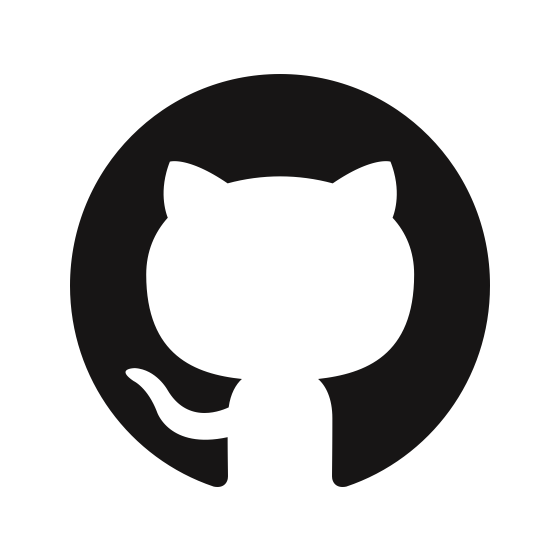}}%
      \hspace{0.45em}%
      \textbf{Code:} %
      \href{https://github.com/SakanaAI/JFinBench}{\nolinkurl{github.com/SakanaAI/JFinBench}}%
      \\[0.3em]
      \raisebox{-0.15em}{\includegraphics[height=1.1em]{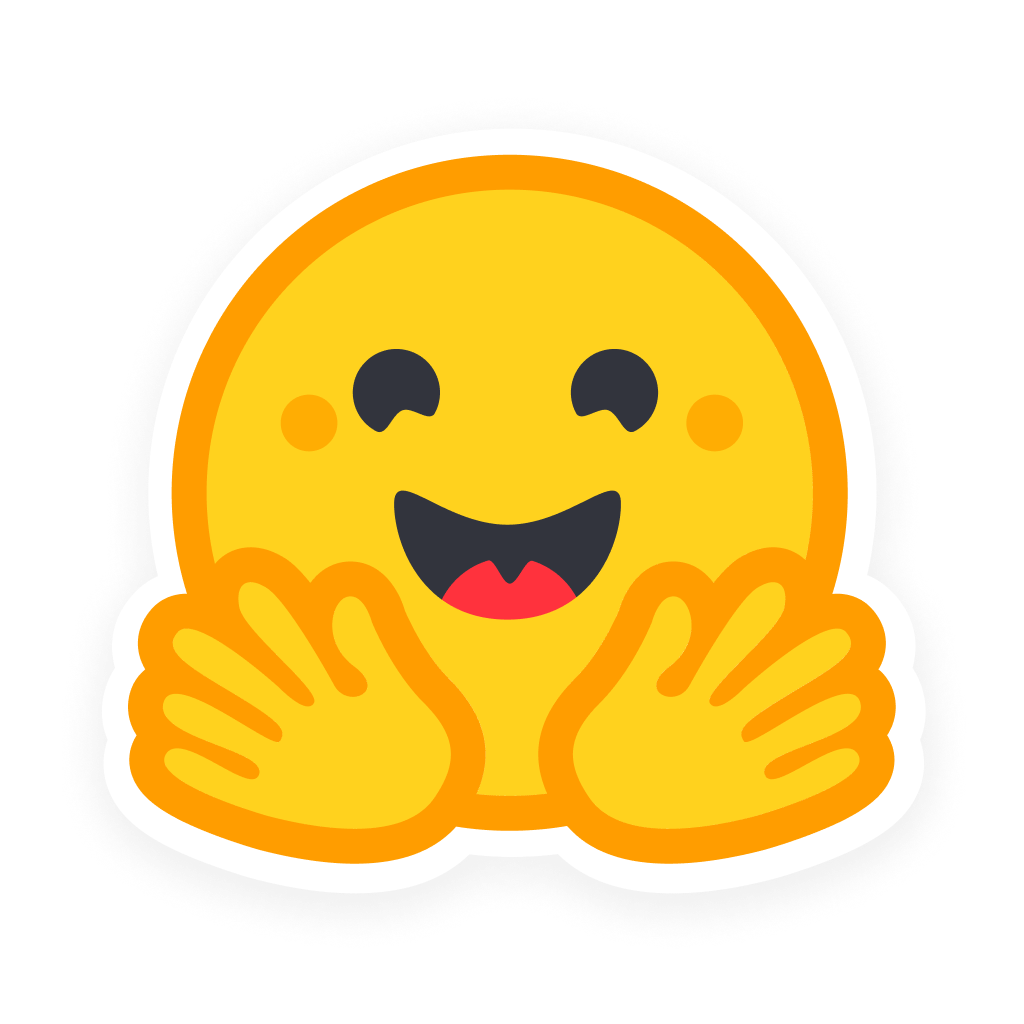}}%
      \hspace{0.45em}%
      \textbf{Data:} %
      \href{https://huggingface.co/datasets/SakanaAI/JFinBench}{\nolinkurl{huggingface.co/datasets/SakanaAI/JFinBench}}%
    \end{tabular}
  \end{center}
  \smallskip
}

\title{\method{}: A Continuously Refreshable Japanese Financial Benchmark\\ from the Latest Certification Exams}

\hypersetup{pdftitle={JFinBench: A Continuously Refreshable Japanese Financial Benchmark from the Latest Certification Exams}, pdfauthor={Issa Sugiura, Taro Makino, Takashi Ishida}}
\author[1,2]{Issa Sugiura}
\author[1]{Taro Makino}
\author[1]{Takashi Ishida}
\affil[1]{Sakana AI}
\affil[2]{Institute of Science Tokyo}

\begin{document}

\begin{abstract}
Large language models (LLMs) are increasingly applied to finance, creating a need for benchmarks that evaluate their capabilities across financial systems with different regulations and practices. However, existing financial benchmarks provide limited coverage of country-specific financial systems, and they are difficult to keep up to date as released questions leak into training data. We introduce JFinBench, a Japanese financial benchmark comprising 321 examples from the 2026 editions of six regularly administered certification exams covering accounting, tax, financial planning, and banking. To construct it, we develop a human-in-the-loop agentic pipeline in which a coding agent writes exam-specific conversion code and a review page, while a human verifies every example and feeds corrections back to the agent. This pipeline adapts to heterogeneous exam formats and reduces the effort of incorporating new exam editions, enabling periodic refreshes that mitigate data contamination. Evaluating 12 recent LLMs, we find that Claude Fable 5 and GPT-6 Astra exceed 93\% overall, while even the best open-weight model, Kimi K3.1 Pro, trails Claude Fable 5 by 12 percentage points. When models can additionally use a terminal and web access, they retrieve the knowledge they lack and every model improves. The gap narrows but about 4 points remain, and the ranking of models stays largely the same, suggesting that knowledge held in the model itself remains valuable even with tools. We release the dataset, the evaluation code, and the construction pipeline.

\end{abstract}
\maketitle

\papercodelink

\begin{figure}[h!]
\centering
\includegraphics[width=\linewidth]{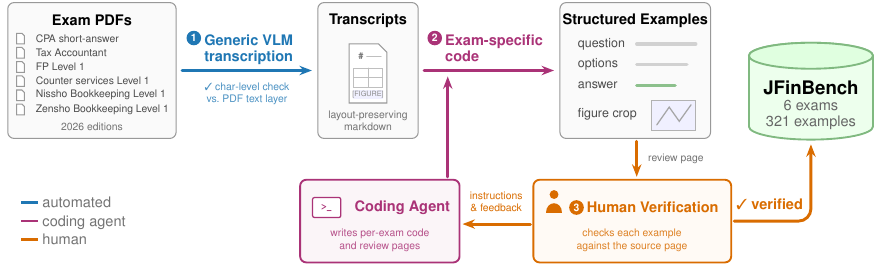}
\caption{Human-in-the-loop agentic construction pipeline of \method{}. Generic scripts transcribe each exam PDF into layout-preserving markdown with a VLM, code written by a coding agent for each exam turns the transcripts into structured examples, and a human verifies every example side by side against the source page and reports problems back to the agent, which fixes its code.}
\label{fig:pipeline}
\end{figure}

\section{Introduction}
\label{sec:intro}

Large language models (LLMs) increasingly take on professional work \citep{massenkoff2026labor}, including software engineering \citep{jimenez2024swebench}, web browsing \citep{wei2025browsecomp}, enterprise workflows \citep{opsahlong2026officeqa,zapier2025automationbench}, economically valuable occupational tasks \citep{patwardhan2025gdpval}, and even running a business \citep{backlund2025vendingbench,sugiura2026coffeebench}. Finance is a critical domain for LLM applications \citep{zhang2026frontierfinance}, spanning everyday tasks such as answering tax and investment questions as well as more complex tasks requiring specialized financial knowledge \citep{sugiura2025edinet,bigeard2025financeagent}. Accurately assessing these capabilities is essential for advancing the application of LLMs in finance.

However, existing financial benchmarks have two important limitations. First, they provide limited coverage of country-specific financial knowledge. Financial expertise depends on local tax rules, accounting standards, and regulations, yet most existing benchmarks \citep{shah2022flue,xie2024finben} focus on English-language data and U.S.-centric settings. For Japanese finance, \citet{hirano2024japanese} built a benchmark from certification exams, but it covers four exams with multiple-choice questions only, leaving free-form answers such as tax calculations and essays, as well as questions that depend on figures, unevaluated. Second, existing benchmarks are difficult to keep up to date. Tax rules and regulations are revised frequently, so questions from past exam editions may test rules that no longer apply. Moreover, once released, benchmark questions can leak into model training data, so evaluation scores may increasingly reflect memorization rather than generalization \citep{sainz2023contamination}. Although regularly incorporating newly released material can mitigate these issues \citep{jain2025livecodebench,white2025livebench,he2026livemathematicianbench}, converting new source documents into reliable structured examples suitable for model evaluation requires substantial effort, hindering frequent updates.

We introduce \method{}, a Japanese financial benchmark comprising 321 examples from the 2026 editions of six financial certification exams. These exams cover diverse areas of Japanese finance and its regulations, from accounting and auditing to tax law, financial planning, and banking. They are also administered regularly, so each new edition supplies fresh expert-written questions that postdate the training data of existing models. To convert the exam PDFs into structured examples suitable for model evaluation, we develop the human-in-the-loop agentic pipeline shown in Figure~\ref{fig:pipeline}. In this pipeline, a coding agent generates exam-specific conversion code and a review page, while a human verifies every example and feeds corrections back to the agent. Because the agent tailors the code to each exam, the pipeline adapts to diverse question formats, including multiple choice, essays, and image-based questions. This reduces the effort required to incorporate new exam editions, making it practical to refresh the benchmark over time.

Evaluating 12 recent LLMs, we find that Claude Fable 5 and GPT-6 Astra exceed 93\% overall and approach saturation on several exams. In contrast, the best open-weight model, Kimi K3.1 Pro, trails Claude Fable 5 by 12 percentage points, suggesting substantial room for improvement in open-weight models' specialized Japanese financial capabilities. This gap stands in contrast to agentic coding, where Kimi K3.1 Pro is reported to outperform GPT-6 Astra on Terminal-Bench \citep{moonshot2026kimik31,merrill2026terminalbench}, suggesting that open-weight models have nearly caught up in domains that model developers prioritize but not in specialized domains such as Japanese finance. Our failure analysis indicates that most errors arise from missing knowledge, while calculation errors are rare. With terminal and web access, most models improve by retrieving missing knowledge from the web. The gap narrows but approximately 4 percentage points remain, and the ranking of models changes little (Spearman $\rho = 0.87$), suggesting that knowledge held in the model's parameters remains valuable even with tools.

\section{Related Work}
\label{sec:related}

\paragraph{Financial benchmarks.}

Financial benchmarks for LLMs have progressed from relatively simple tasks, such as numerical reasoning over corporate filings in FinQA \citep{chen2021finqa}, toward more practical and complex ones. The Finance Agent Benchmark has agents research recent SEC filings with web search and EDGAR access \citep{bigeard2025financeagent}, and FrontierFinance evaluates the full workflow of professional investment research \citep{zhang2026frontierfinance}. These resources primarily focus on English-language financial data and U.S.-centric settings, leaving country-specific financial systems less well covered.

For Japanese finance, where LLM applications are an active area of study \citep{nakagawa2025survey,sugiura2025edinet}, a few benchmarks have begun to address this gap by drawing on certification exams. \citet{masuda2023chatgpt} evaluated ChatGPT on the auditing section of the CPA short-answer exam, and \citet{hirano2024japanese} pioneered benchmark construction from Japanese financial certification exams, covering four exams. \method{} extends this line of work in three directions. First, it draws on the newest exam editions, held in 2026, to limit contamination, whereas prior benchmarks use editions from 2013--2023 and undated sample questions. Second, it covers question formats beyond multiple choice, including image-dependent and free-form questions. Third, it comes with a pipeline that keeps the benchmark refreshable as new editions are released. Table~\ref{tab:related} summarizes the comparison.

\begin{table}[t]
\centering
\small
\setlength{\tabcolsep}{3pt}
\begin{tabular}{@{}lccccc@{}}
\toprule
\multirow{2}{*}{Benchmark} & \multirow{2}{*}{Exams} & \multicolumn{3}{c}{Question types} & \multirow{2}{*}{Latest ed.} \\
\cmidrule(lr){3-5}
 & & MC & Free & VQA & \\
\midrule
\citet{masuda2023chatgpt} & 1 & \cmark & \xmark & \xmark & \xmark \\
\citet{hirano2024japanese} & 4 & \cmark & \xmark & \xmark & \xmark \\
\method{} (ours) & 6 & \cmark & \cmark & \cmark & \cmark \\
\bottomrule
\end{tabular}
\caption{Exam-based Japanese financial benchmarks. MC: multiple choice. Free: free-response, form, and essay answers graded with rubrics. VQA: image-dependent questions retained as visual inputs.}
\label{tab:related}
\end{table}

\paragraph{Exam-based benchmarks.}

Examinations provide a natural source of evaluation data because they are authored by domain experts and designed to distinguish different levels of human competence: MMLU \citep{hendrycks2021mmlu} draws on school-level examinations and standardized admission tests. More recent studies have turned to professional licensing examinations, including the bar exam \citep{openai2023gpt4,katz2024bar,choi2023chatgpt,iu2023chatgpt} and medical licensing exams \citep{nori2023capabilities,kasai2023igakuqa,xun2026jmed48k}, and the mathematics competition AIME \citep{maa_aime} has become a standard benchmark for reasoning models. \method{} follows this line of work by using Japanese financial certification exams as a source of expert-authored evaluation material.

\paragraph{Live benchmarks.}

A persistent challenge for benchmark-based evaluation is data contamination: benchmark questions may enter model training data, so scores come to reflect memorization rather than generalization \citep{sainz2023contamination}. Post-hoc detection methods exist \citep{oren2024proving,deng2024investigating}, but a more fundamental remedy is to evaluate on material that postdates the training data, as in live benchmarks such as LiveCodeBench, which periodically incorporate newly released problems \citep{jain2025livecodebench,white2025livebench,he2026livemathematicianbench}. \method{} brings this live-benchmark approach to Japanese finance, using the certification exams, which are held regularly and published officially, as a recurring source of fresh, expert-authored evaluation material.

\paragraph{Scalable benchmark construction.}

Building a benchmark by hand is laborious \citep{merrill2026terminalbench}, so a growing line of work uses programs and agents to make its construction more efficient and scalable. One approach mines existing artifacts and converts them automatically: SWE-bench \citep{jimenez2024swebench} builds tasks from real GitHub issues and the pull requests that resolved them, and ProgramBench \citep{yang2026programbench} uses existing programs as references and generates behavioral tests by agent-driven fuzzing. Agents increasingly author the questions as well: OfficeQA Pro V2 \citep{databricks2026officeqav2} generates them with synthesis, solver, and verification agents and keeps humans only for the final review. \method{} likewise builds on existing artifacts, namely the official exam PDFs, and makes their conversion scalable with a human-in-the-loop agentic pipeline, in which a coding agent automates most of the work and a human verifies the results.

\section{\method{}}
\label{sec:bench}

\method{} is a benchmark of Japanese financial capabilities comprising 321 examples drawn from six certification exams held between May and August 2026, as summarized in Table~\ref{tab:stats}.

\begin{table*}[t]
\centering
\footnotesize
\setlength{\tabcolsep}{4pt}
\begin{tabular}{@{}llrrrrrrr@{}}
\toprule
\multirow{2}{*}{Exam} & \multirow{2}{*}{Exam date} & \multirow{2}{*}{Sections} & \multicolumn{4}{c}{Question type} & \multirow{2}{*}{Examples} & \multirow{2}{*}{w/ image} \\
\cmidrule(lr){4-7}
 & & & MCQ & Short & Form & Essay & & \\
\midrule
CPA short-answer exam & May 24, 2026 & 4 & 93 & -- & -- & -- & 93 & -- \\
Certified Public Tax Accountant & Aug 4--6, 2026 & 11 & -- & -- & 17 & 23 & 40 & 1 \\
FP Level 1, written & May 24, 2026 & 2 & 50 & 15 & -- & -- & 65 & 6 \\
Financial counter services Level 1 & May 2026$^{*}$ & 2 & 50 & 16 & -- & -- & 66 & 5 \\
Nissho Bookkeeping Level 1 & Jun 14, 2026 & 4 & -- & 23 & 1 & -- & 24 & -- \\
Zensho Bookkeeping Level 1 & Jun 28, 2026 & 2 & -- & 33 & -- & -- & 33 & -- \\
\midrule
Total & & 25 & 193 & 87 & 18 & 23 & 321 & 12 \\
\bottomrule
\end{tabular}
\caption{Statistics of \method{}. Examples is the number of separately graded units, as defined in Section~\ref{sec:stats}. Question types: MCQ (multiple choice), Short (short answer), Form (structured answer forms), and Essay. The w/ image column counts examples whose question includes an image. $^{*}$Administered by computer-based testing (CBT) without a fixed exam day. We use the question set published in May 2026.}
\label{tab:stats}
\end{table*}

\subsection{Source exams}
\label{sec:exams}

We select exams that cover a core area of Japanese financial practice, whose questions are published as official PDFs by the administering organization with publicly available answer keys, and whose most recent edition was held in 2026. When an exam is offered at multiple levels, we use the highest, Level~1. Below, we give an overview of each of the six exams.\footnote{The official Japanese names and licenses of the selected exams are given in Appendix~\ref{app:examnames}.}

\paragraph{CPA short-answer exam.} The first-stage screening exam of Japan's national CPA qualification, administered by the Certified Public Accountants and Auditing Oversight Board under the Financial Services Agency. The exam consists of four subjects, namely financial accounting, management accounting, auditing, and corporate law, all answered as multiple-choice questions (MCQs).

\paragraph{Certified Public Tax Accountant exam.} The national licensing exam for tax practitioners, administered by the National Tax Council under the National Tax Agency. The exam consists of 11 subjects, two in accounting and nine in tax law, all answered as essays or long-form calculations.

\paragraph{FP skill test.} The national certification for financial planners at the practitioner level, consisting of a written test and a practical test. We include the written test. It comprises a basic part, which asks knowledge MCQs on topics such as pensions, insurance, tax planning, and inheritance, and an applied part, which poses client scenarios whose answers, such as pension benefits or inheritance-tax amounts for a given household, must be computed.

\paragraph{Financial counter services skill test.} The national certification for bank-counter professionals, combining knowledge MCQs with counter consultations answered as MCQs and short responses.

\paragraph{Nissho Bookkeeping.} Japan's de facto standard bookkeeping certification. Level~1 covers advanced financial and cost accounting, such as consolidated statements and standard costing, answered mostly as short computed values, with one full answer form.

\paragraph{Zensho Bookkeeping.} A bookkeeping test administered by the National Association of Commercial High Schools and taken mainly by commercial high school students, covering financial and cost accounting answered mostly as short computed values.

\subsection{Construction pipeline}
\label{sec:pipeline}

The exam PDFs differ in layout across exams and often across questions within an exam, and their tables, answer forms, and figures do not survive standard text extraction tools such as PyMuPDF \citep{pymupdf}. We therefore build \method{} with the human-in-the-loop agentic pipeline shown in Figure~\ref{fig:pipeline}, which converts the PDFs into structured, machine-readable examples in three steps: generic scripts shared by all exams transcribe the pages, a coding agent writes exam-specific code that turns the transcripts into examples, and a human verifies every example and feeds corrections back to the agent. We describe each step below.

\paragraph{Generic transcription.} Shared scripts render each page as an image at 300\,dpi and transcribe it into markdown with a VLM, Gemini 3.8 Flash \citep{google2026gemini38}, which preserves the page layout: tables become markdown tables, and figures that cannot be transcribed as text are replaced by explicit placeholders rather than silently dropped. Each transcript is compared with the PyMuPDF-extracted text as character multisets, flagging omissions and hallucinations for inspection.

\paragraph{Exam-specific conversion by a coding agent.} For each exam, a coding agent, Claude Code \citep{anthropic2025claudecode}, writes the code that splits the transcripts into typed example records, attaches options and answer keys, crops figure regions for visual examples, removes non-content artifacts such as page headers and a preparatory school's predicted point allocations, and generates a side-by-side review page, shown in Appendix~\ref{app:review}. The agent works from natural-language instructions that specify the source URLs and the statutory basis date, so adding an exam requires no manual programming.

\paragraph{Human verification and feedback.} For every example, one of the authors, a native Japanese speaker, compares the final record against the source page on this review page. The reviewer reports problems in natural language, ranging from misparsed options and leftover pagination notes to missing reference material and unclear answer formats, and the agent fixes its code or the records and regenerates the affected examples for re-verification. All 321 examples were verified this way, which took roughly two hours per exam, or about 12 hours in total. The six exams differ widely in format, from multiple choice to blank-filling bookkeeping forms and tax essays across 11 subjects, yet share the same generic scripts, so adding a new exam or edition requires only new agent-written code and human verification. Since neither the generic scripts nor the agent-driven workflow is specific to finance, the same pipeline could be applied to other certification exams, such as the national medical licensing or bar exams.

\subsection{Statistics and question types}
\label{sec:stats}

Table~\ref{tab:stats} breaks the 321 examples down by exam and question type. \method{} has four question types, each with its own grading method described in Section~\ref{sec:scoring}. \emph{MCQ} examples are answered by selecting an option. \emph{Short answer} examples ask for a term or an amount written directly. \emph{Form} examples require a structured answer form with up to dozens of interdependent blanks, graded blank by blank. \emph{Essay} examples ask for a long-form theory answer, graded with a rubric derived from the answer key. The 321 examples, each a separately graded unit, derive from 267 source problems. Multiple choice accounts for 193 examples. The remaining 128 require generated answers, from one-line values to full essays. Twelve examples additionally include images, such as land-plot diagrams, served to the models as visual inputs. Representative examples are shown in Appendix~\ref{app:examples}.

\section{Experimental Setup}
\label{sec:setup}

We evaluate models primarily without tools, answering each example in a single API call, so that scores reflect the knowledge stored in the model's parameters and its reasoning. Section~\ref{sec:agentic} additionally evaluates a setting with tools, where models can use a terminal and web access to run computations and retrieve knowledge they lack.

\subsection{Models}
We evaluate 12 models: Claude Fable 5 \citep{anthropic2026fable}, GPT-6 Astra \citep{openai2026gpt6}, and Gemini 3.8 Flash \citep{google2026gemini38} as proprietary models, and Kimi K3.1 Pro, K3, and K2.6 \citep{moonshot2026kimik31,moonshot2026kimik3,moonshot2026kimik26}, DeepSeek V4.1 Flash 1017 \citep{deepseek2026v4,deepseek2026v41flash1017}, Qwen3.8 Flash \citep{alibaba2026qwen38flash}, MiMo-V2.6-Pro and MiMo-V2.6-Flash \citep{xiaomi2026mimo}, GLM 5.3 Flash \citep{zai2026glm5}, and Mistral Large 4 \citep{mistral2026large4} as open-weight models. All models accept image inputs. Gemini 3.8 Flash is accessed through Google's official API. All other models are accessed through the OpenRouter API,\footnote{\url{https://openrouter.ai}} pinned to the model developer's official provider, except for DeepSeek V4.1 Flash 1017, which is served by DeepInfra. Release dates and reported knowledge cutoffs are listed in Appendix~\ref{app:cutoffs}, and every exam postdates every reported cutoff.

\subsection{Inference settings}
We use the developer's recommended sampling parameters where published. We set the maximum output length to 131{,}072 tokens. For each model, we run the maximum and the minimum reasoning effort it supports. Models without an effective effort control use a single setting with thinking enabled. Because neither model outputs nor judge decisions are deterministic, every combination of model and effort level is run three times, and we report the mean.

\subsection{Scoring}
\label{sec:scoring}
Each example is scored by a method matched to its question type.
\begin{itemize}\setlength{\itemsep}{0pt}
\item \textbf{MCQ}: regex-based answer extraction, with an LLM-judge fallback \citep{zheng2023judging} for malformed outputs.
\item \textbf{Short answer}: an equivalence judge that accepts mathematically or legally equivalent phrasings.
\item \textbf{Form}: blank-level partial credit, i.e., the fraction of blanks matching the answer key, with up to about 40 blanks per form.
\item \textbf{Essay}: rubric-based grading, where the key points of the answer key serve as the rubric and the score is the fraction of key points the response covers.
\end{itemize}
\noindent All judges use the same grader model, GPT-5.1 \citep{openai2025gpt51}. The full judge prompts are in Appendix~\ref{app:graders}, and the sensitivity of scores to the judge model is analyzed in Appendix~\ref{app:judge}, where re-grading with a second judge, Claude Opus 5.5 \citep{anthropic2026opus55}, yields closely matching scores. An exam's score is the mean over its examples, and the overall score is the mean over the six exams with equal weight. Cost is computed from measured input/output tokens at the official list prices given in Appendix~\ref{app:prices} and excludes the grader.

\subsection{Prompts}
Every prompt states the exam context together with the exam date and the statutory basis date, for example that ``laws in force as of April 1, 2026 apply,'' because the correct answer to many questions can change with revisions of statutes, regulations, and accounting standards. Models are thus graded under the law in force at exam time. Answer-format instructions are specific to the question type, and the full MCQ template is in Appendix~\ref{app:prompt}.

\section{Results}
\label{sec:results}

\subsection{Main results}
\label{sec:mainresults}

\begin{table*}[t]
\centering
\small
\setlength{\tabcolsep}{5pt}
\resizebox{\textwidth}{!}{%
\begin{tabular}{@{}lrrrrrr@{\hspace{10pt}}rrr@{}}
\toprule
& \multicolumn{6}{c}{Score by exam (\%)} & \multicolumn{3}{c}{Overall} \\
\cmidrule(r{4pt}){2-7}\cmidrule(l){8-10}
Model & CPA & Tax & FP & Counter & Nissho & Zensho & \textbf{Score (\%)} & Cost (\$/Q) & Tok.\ (K/Q) \\
\midrule
\rowcolor{gray!15}\multicolumn{10}{@{}l}{\emph{Proprietary models}} \\
Claude Fable 5 (max)      & \textbf{97.1} & \textbf{79.4} & \textbf{100.0} & \textbf{100.0} & \textbf{100.0} & \textbf{98.0} & \textbf{95.8}{\scriptsize$\pm$0.4} & 0.555 & 10.9 \\
GPT-6 Astra (max)         & 96.8 & 73.7 & 99.5 & 99.0 & 96.9 & 96.0 & 93.6{\scriptsize$\pm$0.6} & 0.248 & 4.8 \\
Gemini 3.8 Flash (high)   & 92.5 & 60.7 & 96.9 & 98.5 & 98.2 & 85.9 & 88.8{\scriptsize$\pm$1.3} & 0.025 & 6.6 \\
\midrule
\rowcolor{gray!15}\multicolumn{10}{@{}l}{\emph{Open-weight models}} \\
Kimi K3.1 Pro (max)       & 84.2 & 48.1 & 87.7 & 90.9 & 96.8 & 92.9 & 83.4{\scriptsize$\pm$0.9} & 0.145 & 14.3 \\
Kimi K3 (max)             & 78.9 & 48.3 & 81.0 & 85.9 & 92.6 & 92.9 & 79.9{\scriptsize$\pm$2.3} & 0.253 & 16.6 \\
DS V4.1 Flash 1017 (max)  & 73.1 & 39.9 & 66.2 & 80.8 & 89.9 & 88.9 & 73.1{\scriptsize$\pm$0.7} & 0.007 & 23.5 \\
Qwen3.8 Flash (thinking)  & 65.9 & 28.8 & 39.5 & 73.7 & 87.4 & 79.8 & 62.5{\scriptsize$\pm$1.0} & 0.020 & 43.2 \\
MiMo-V2.6-Pro (thinking)  & 63.4 & 23.1 & 50.3 & 76.3 & 76.4 & 82.8 & 62.1{\scriptsize$\pm$1.0} & 0.030 & 33.9 \\
Kimi K2.6 (thinking)      & 54.5 & 25.2 & 42.1 & 74.8 & 84.6 & 73.7 & 59.1{\scriptsize$\pm$0.5} & 0.058 & 14.2 \\
Mistral Large 4 (thinking)& 54.8 & 25.5 & 44.1 & 76.3 & 74.8 & 73.7 & 58.2{\scriptsize$\pm$2.3} & 0.061 & 29.0 \\
GLM 5.3 Flash (max)       & 56.6 & 23.6 & 40.0 & 65.7 & 80.4 & 79.8 & 57.7{\scriptsize$\pm$2.3} & 0.013 & 26.4 \\
MiMo-V2.6-Flash (thinking)& 50.5 & 18.5 & 36.4 & 62.1 & 65.3 & 72.7 & 50.9{\scriptsize$\pm$1.0} & 0.010 & 34.4 \\
\bottomrule
\end{tabular}%
}
\caption{Model performance on \method{} (w/o tools) at maximum reasoning effort (mean over 3 runs, with the standard deviation for the overall score). The best open-weight model trails the best proprietary model by 12 points. Full results with standard deviations are in Appendix~\ref{app:low}.}
\label{tab:main}
\end{table*}

Table~\ref{tab:main} reports the main results without tools at maximum reasoning effort. Claude Fable 5 and GPT-6 Astra achieve the highest overall scores of 95.8 and 93.6. The two leaders score 96--100 on every exam except the tax accountant exam, where the best score is 79.4. The three proprietary models occupy the top three ranks. The strongest open-weight model, Kimi K3.1 Pro, reaches an overall score of 83.4, a full 12 percentage points behind Claude Fable 5, and every open-weight model scores below every proprietary one. This gap contrasts with agentic coding, a capability that model developers actively prioritize, where Kimi K3.1 Pro is reported to score 65.7\% on Terminal-Bench 4.0, above GPT-6 Astra's 58.2\% \citep{moonshot2026kimik31}. On \method{}, GPT-6 Astra instead leads Kimi K3.1 Pro by 10.2 points. This clear separation suggests that open-weight models lack much of the specialized knowledge of Japanese finance that proprietary models have acquired.

\subsection{Performance vs.\ cost}

\begin{figure*}[t]
  \centering
  \includegraphics[width=0.72\textwidth]{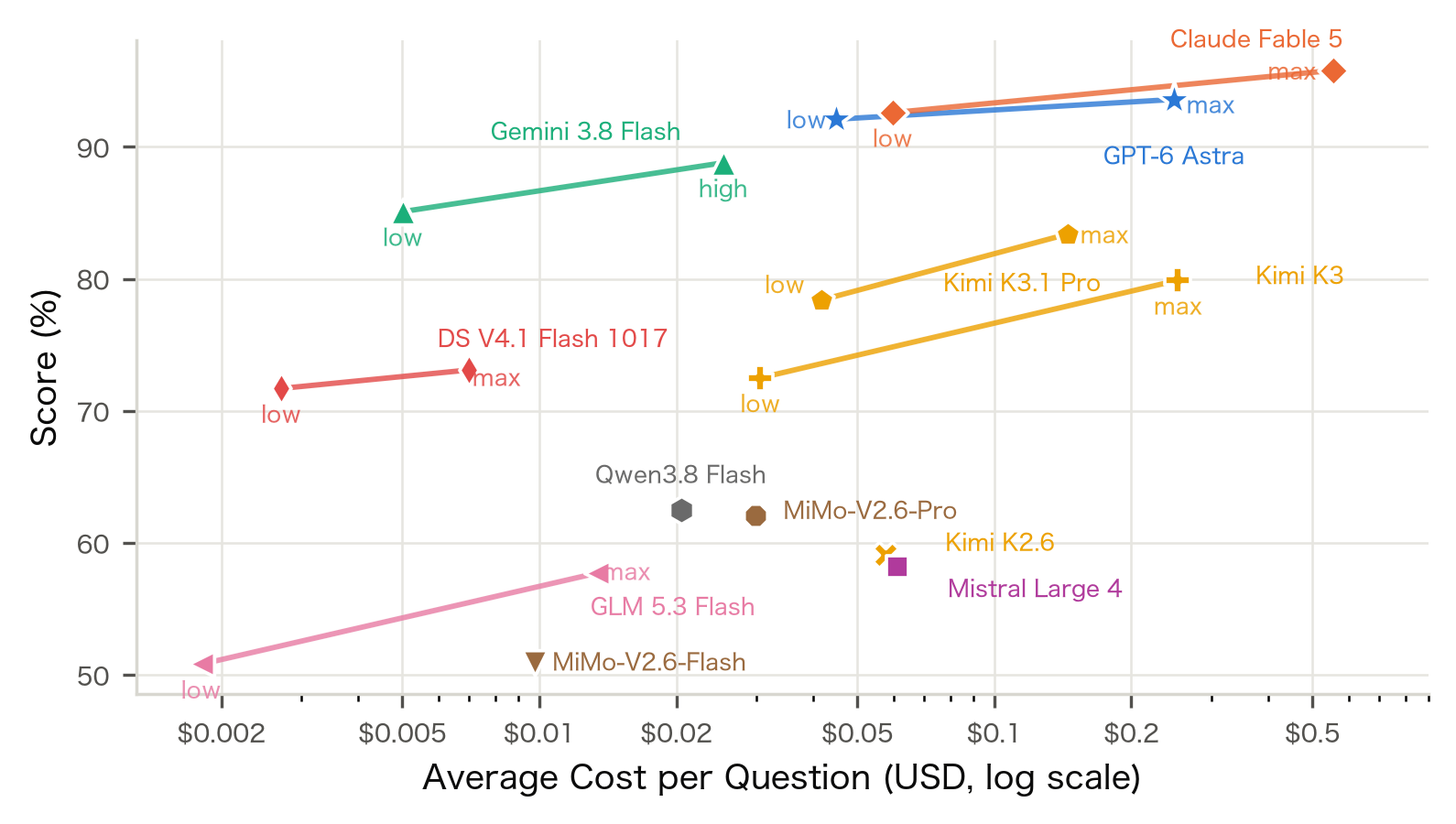}
  \caption{Model performance vs.\ cost (w/o tools). For every model with an effort setting, raising the reasoning effort increases both cost and score, showing a trade-off between the two.}
  \label{fig:cost}
\end{figure*}
Reasoning models \citep{openai2024o1,guo2025deepseekr1} expose settings, such as reasoning effort, that control how much test-time compute they spend, so models should also be compared on the cost at which they reach a given score \citep{brown2026ttc}. Figure~\ref{fig:cost} therefore plots overall score against API cost per question, at the lowest and the maximum reasoning-effort settings. For every model, raising the effort increases both the score and the cost. Gemini 3.8 Flash stands out for its low cost, outscoring every open-weight model even at low effort. Appendix~\ref{app:low} additionally reports the full low-effort results and plots score against output tokens.

\begin{figure}[t]
\centering
\begin{minipage}[b]{0.40\textwidth}
\centering
    \includegraphics[width=\linewidth]{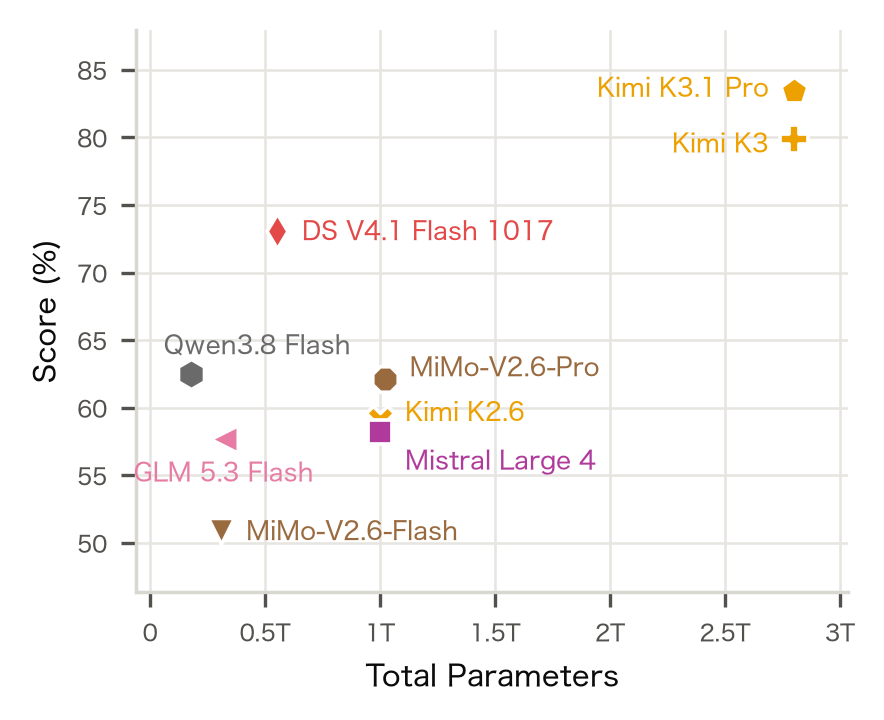}
  \caption{Overall score (w/o tools) versus total parameter count for the open-weight models (maximum effort). Score correlates positively with parameter count (Pearson $r = 0.82$).}
  \label{fig:params}
\end{minipage}\hfill
\begin{minipage}[b]{0.56\textwidth}
\centering
\includegraphics[width=\linewidth]{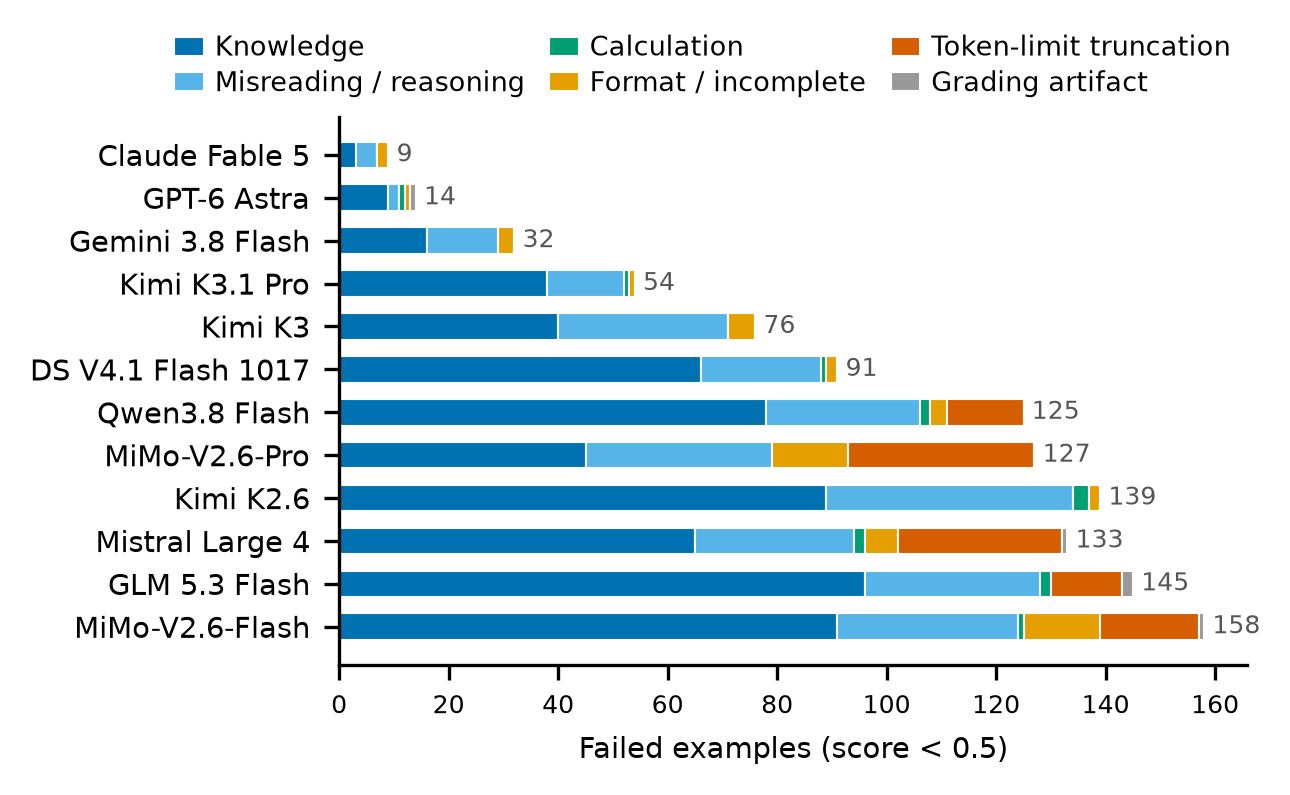}
\caption{Error type distribution of each model (w/o tools, failed examples with score $<$ 0.5, run 1 at maximum effort). Knowledge errors are the most common cause for most models, and token-limit truncation is concentrated in a handful of models.}
\label{fig:failures}
\end{minipage}
\end{figure}

\begin{table}[t]
\centering
\small
\setlength{\tabcolsep}{8pt}
\begin{tabular}{@{}lrlr}
\toprule
Model & w/o tools & w/ tools (harness) & $\Delta$ \\
\midrule
Claude Fable 5 (max)       & 95.8 & 96.4 (Terminus-2) & \cellcolor{deltagreen!1}$+$0.6 \\
GPT-6 Astra (max)          & 93.6 & 97.3 (Terminus-2) & \cellcolor{deltagreen!5}$+$3.7 \\
Gemini 3.8 Flash (high)    & 88.8 & 94.2 (Terminus-2) & \cellcolor{deltagreen!7}$+$5.4 \\
\midrule
Kimi K3.1 Pro (max)        & 83.4 & 92.9 (Kimi Code) & \cellcolor{deltagreen!13}$+$9.5 \\
Kimi K3 (max)              & 79.9 & 91.0 (Kimi Code) & \cellcolor{deltagreen!15}$+$11.1 \\
DS V4.1 Flash 1017 (max)   & 73.1 & 91.5 (mini-SWE-agent) & \cellcolor{deltagreen!25}$+$18.4 \\
Qwen3.8 Flash (thinking)   & 62.5 & 85.9 (mini-SWE-agent) & \cellcolor{deltagreen!31}$+$23.4 \\
MiMo-V2.6-Pro (thinking)   & 62.1 & 86.3 (mini-SWE-agent) & \cellcolor{deltagreen!32}$+$24.2 \\
Kimi K2.6 (thinking)       & 59.1 & 74.7 (mini-SWE-agent) & \cellcolor{deltagreen!21}$+$15.6 \\
Mistral Large 4 (thinking) & 58.2 & 75.1 (Mistral Vibe) & \cellcolor{deltagreen!23}$+$16.9 \\
GLM 5.3 Flash (max)        & 57.7 & 87.6 (Terminus-2) & \cellcolor{deltagreen!40}$+$29.9 \\
MiMo-V2.6-Flash (thinking) & 50.9 & 75.8 (MiMo Code) & \cellcolor{deltagreen!33}$+$24.9 \\
\bottomrule
\end{tabular}
\caption{Comparison of each model's score without and with tools (mean over 3 runs). For w/ tools, we report the best score among the Terminus-2, mini-SWE-agent, and official harnesses (Table~\ref{tab:harness}). With shell and web access, every model improves, and the open-weight models gain 9.5 to 29.9 points.}
\label{tab:agentic}
\end{table}

\begin{figure}[t]
\begin{minipage}[b]{0.43\textwidth}
\centering
\includegraphics[width=\linewidth]{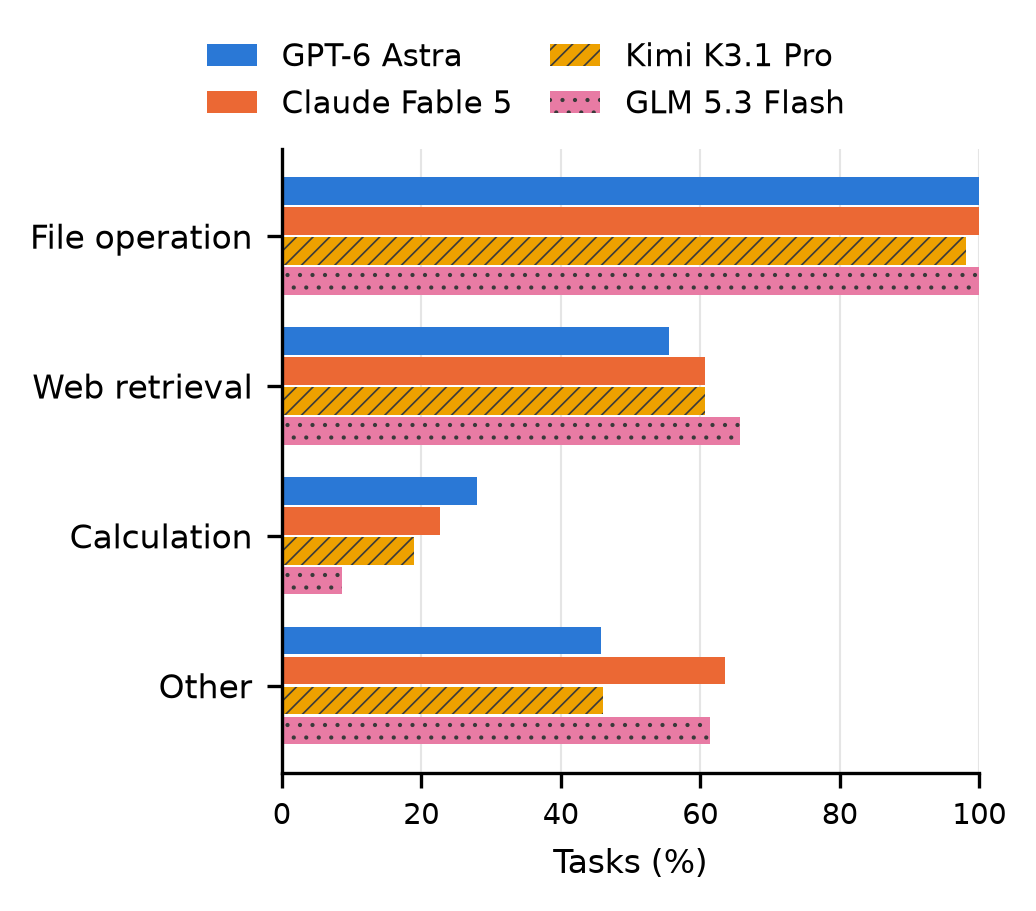}
\end{minipage}\hfill
\begin{minipage}[b]{0.54\textwidth}
\centering
\includegraphics[width=\linewidth]{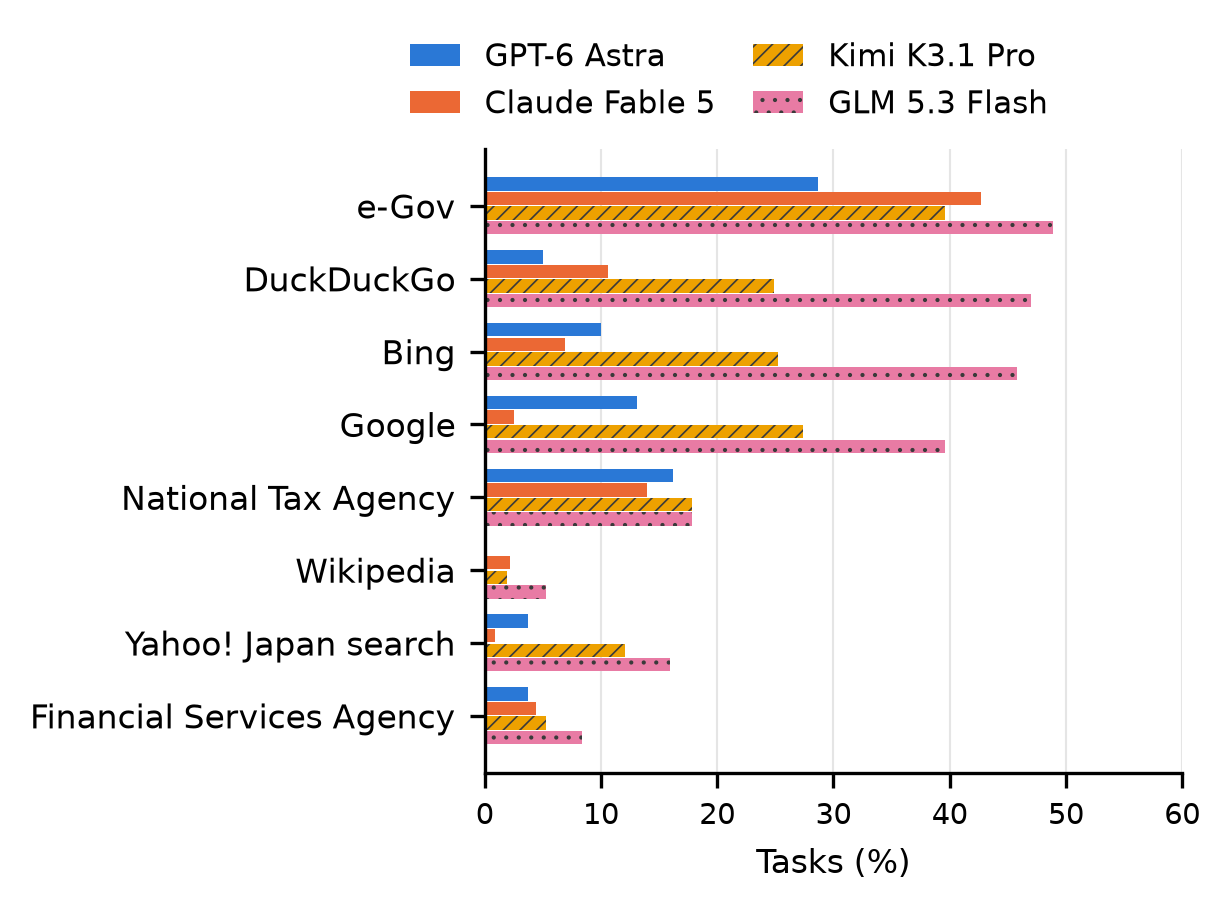}
\end{minipage}

\begin{minipage}[t]{0.43\textwidth}
\caption{Breakdown of the main tools that four representative models used in the with-tools evaluation (Terminus-2, run 1).}
\label{fig:agentic_tools}
\end{minipage}\hfill
\begin{minipage}[t]{0.54\textwidth}
\caption{Breakdown of the domains that four representative models accessed most often in the with-tools evaluation (Terminus-2, run 1).}
\label{fig:agentic_sites}
\end{minipage}
\end{figure}

\subsection{Performance vs.\ parameter count}
Model scale is considered an important factor underlying performance on knowledge-intensive benchmarks \citep{geva2021kv,roberts2020knowledge,allenzhu2024physics}. Figure~\ref{fig:params} plots overall score against total rather than active parameter count for the nine open-weight models, whose parameter counts are public. Score generally increases with parameter count (Pearson $r = 0.82$), but parameter count is not the only factor. Kimi K3.1 Pro, post-trained from Kimi K3, improves on it by 3.5 points at the same 2.8T parameters, which points to the effect of post-training, and DeepSeek V4.1 Flash 1017 outscores the trillion-parameter Kimi K2.6, MiMo-V2.6-Pro, and Mistral Large 4 with about half their parameters.

\subsection{Failure analysis}
\label{sec:failure}
To understand why models fail, we classify every failed example (score $<$ 0.5) from the first maximum-effort run of each evaluated model into one of six primary causes.
\begin{itemize}\setlength{\itemsep}{0pt}
\item \textbf{Knowledge}: the response lacks or misstates the fact or rule the question tests.
\item \textbf{Misreading / reasoning}: the response misreads the question or draws a wrong conclusion from correct knowledge.
\item \textbf{Calculation}: arithmetic errors alone.
\item \textbf{Format / incomplete}: the answer deviates from the specified format, is missing required parts, or is unparseable.
\item \textbf{Token-limit truncation}: generation, including reasoning tokens, hits the output-token limit with no final answer.
\item \textbf{Grading artifact}: the response is equivalent to the answer key but was scored as wrong.
\end{itemize}

\noindent Token-limit truncation is detected automatically. The other categories are assigned by an LLM classifier (GPT-5.1), following the LLM-judged failure analysis of Terminal-Bench \citep{merrill2026terminalbench}. The classifier reads the question, the answer key, and the model's response, and picks the first error that derails the response. To validate the classifier, one of the authors independently labeled 30 sampled failures, five per category, including automatically detected token-limit truncation. The labels agree on 26 of 30 (Cohen's $\kappa = 0.84$), and all four disagreements are between the knowledge and misreading/reasoning categories. Figure~\ref{fig:failures} shows the distribution.

\paragraph{Knowledge errors dominate.} Knowledge is the most common primary cause overall. A recurring case is statute recall, where models state the purpose of a provision correctly but cannot reproduce its enumerated requirements.

\paragraph{Overlong reasoning hits the token limit.} Token-limit truncation is concentrated in five open-weight models, the MiMo-V2.6 models, Mistral Large 4, Qwen3.8 Flash, and GLM 5.3 Flash. These models sometimes fail to control the length of their reasoning and keep generating until they hit the output-token limit.

\subsection{Evaluation with tools}
\label{sec:agentic}

Results without tools measure parametric knowledge and in-context reasoning, but real financial work increasingly involves tools \citep{bigeard2025financeagent}. We therefore also evaluate every model in a setting \emph{with tools}, where for each question the model can use a terminal and web access over multiple turns before answering. We run this evaluation with Harbor \citep{shi2026harboradaptersharborindexinfrastructure}, a framework for agentic evaluation that runs each task in an isolated container. Each model is evaluated with Harbor's Terminus-2 agent harness, with mini-SWE-agent \citep{yang2024sweagent}, and with the developer's official harness where one exists, and we report the score of its best harness. Because the web may contain the published answers, the instructions prohibit consulting answer keys, and we audit every trajectory with an LLM. Attempts that used answer material, 8 of the 11{,}556 with Terminus-2 (0.07\%), score 0 (Appendix~\ref{app:agentic_setup}). Appendix~\ref{app:agentic} describes the setup in detail and reports further analyses.

\paragraph{Tools raise every model.} Table~\ref{tab:agentic} compares the two settings. Tool access raises the score of every model, and most for GLM 5.3 Flash, which gains 29.9 points. Taking each model's best harness, the gap between the best open-weight and the best proprietary model narrows from 12.4 to 4.4 points, with Kimi K3.1 Pro reaching 92.9 with Kimi Code against GPT-6 Astra's 97.3 with Terminus-2. Yet tools do not erase the differences between models. The ranking changes little between the two settings (Spearman $\rho = 0.87$ over the 12 models), and the same three models lead in both. The main exception is GLM 5.3 Flash, which rises from 11th to 7th. Models that know more without tools thus also score higher with them, suggesting that knowledge stored in the parameters helps an agent decide what to look up and how to apply what it finds. Retrieval complements rather than replaces this knowledge, so \method{} remains informative when models can use tools. Appendix~\ref{app:harness} compares the harnesses in detail.

\paragraph{How agents use tools.} To see how the models improve with tools, we analyze which kinds of tools the agents use in the first run, where an attempt is one agent session on one question. GPT-6 Luna \citep{openai2026gpt6sol} reads all commands of each attempt and labels whether it involves file operation, web retrieval, calculation, or other commands. Figure~\ref{fig:agentic_tools} shows, for two proprietary and two open-weight models with Terminus-2, the one harness common to all models, the share of the 321 tasks in which the agent uses each kind of tool. Every model operates on files in almost every task, which is expected because the final answer must be written to a file. Web retrieval occurs in 56 to 66\% of the tasks, as the agents look up knowledge they lack. Calculation occurs in 9 to 28\% of the tasks, where the agents compute amounts such as taxes, depreciation, and present values with Python or \texttt{bc}.

\paragraph{Agents consult the statutes.} Figure~\ref{fig:agentic_sites} shows the share of tasks in which the agent accesses each of the most frequent domains. All four models most often consult e-Gov, the government's statute database,\footnote{\url{https://laws.e-gov.go.jp/}} and the open-weight models also query search engines far more often, for example DuckDuckGo in 47\% of the tasks for GLM 5.3 Flash against at most 11\% for the proprietary models. Figure~\ref{fig:agentic_traj} in Appendix~\ref{app:agentic_example} shows a trajectory in which the agent retrieves the statutes in force on the exam's basis date from e-Gov and answers a question it fails without tools. These results suggest that, with tools, models raise their scores mainly by supplementing missing knowledge and, on computation questions, by calculating exactly.

\section{Conclusion}
\label{sec:conclusion}

We introduced \method{}, a Japanese financial benchmark built from the 2026 editions of six certification exams with a human-in-the-loop agentic pipeline that reduces the human effort of future updates. The leading proprietary models are near the ceiling except on tax practice, while the best open-weight model trails the best proprietary model by 12 points. Missing knowledge is the main challenge. Tool use helps substantially, raising every model and narrowing but not closing the gap between the best proprietary and best open-weight models, from 12 to about 4 points, while the ranking of models stays largely the same. We hope \method{} encourages progress in domains that general benchmarks overlook.

\section*{Limitations}
\label{sec:limitations}

\paragraph{Coverage.} We include only exams whose questions are published as PDFs by the administering organization. Exams that do not publish past questions as official PDFs, such as the Securities Analyst (CMA) exam,\footnote{\url{https://www.saa.or.jp/}} are out of scope.
\paragraph{Contamination.} Every exam postdates every reported knowledge cutoff listed in Appendix~\ref{app:cutoffs}, but this temporal separation does not guarantee the absence of contamination. Most vendors do not report a cutoff at all, and reported cutoffs are self-declared, vary in definition across developers, and may differ from the effective cutoff of the training data \citep{cheng2024dated}. Models may also be post-trained on data collected after the stated date. Refreshing the benchmark with future exam editions remains the primary mitigation.
\paragraph{Saturation.} The two leading proprietary models already score 96--100 on every exam except the tax accountant exam, so \method{} offers limited headroom for differentiating them. The benchmark remains well suited to tracking the progress of open-weight models, which still trail by 12 percentage points. For frontier models, promising directions include filtering future editions to the hardest questions, as in Humanity's Last Exam \citep{phan2025hle}, commissioning new expert-authored questions \citep{merrill2026terminalbench}, and building agentic financial benchmarks \citep{bigeard2025financeagent,zhang2026frontierfinance}.

\section*{Ethical Considerations}

The exam material contains no personal data, as persons appearing in problem statements are fictitious. A high \method{} score does not constitute a professional qualification, and our results should not be read as evidence that current models can substitute for licensed tax accountants or CPAs. Because \method{} is built from officially published examinations, we see little risk of social harm or misuse arising from the benchmark itself.

\section*{Acknowledgments}
We thank Alain Saas for helpful discussions.

\bibliographystyle{acl_natbib}
\bibliography{custom}

\appendix

\section{Benchmark Details}
\label{app:bench}

\subsection{Exam Names, Sources, and Licenses}
\label{app:examnames}

Table~\ref{tab:licenses} lists the official Japanese name of each exam, the source and license of its questions and answer keys, and how we distribute them. The questions and answer keys come from the publishers' official pages, except that the answer keys of the Certified Public Tax Accountant exam and Nissho Bookkeeping are the answer bulletins of the preparatory school TAC. Copyright, where applicable, remains with the respective rights holders, and the hosted materials are provided under the source-specific terms in the table. \method{} was prepared by the authors from the publishers' PDFs and is not an official publication of, or endorsed by, the examination bodies or answer providers. Our code is released under the Apache License 2.0.

\begin{table*}[t]
\centering
\footnotesize
\setlength{\tabcolsep}{4pt}
\begin{tabular}{@{}>{\raggedright\arraybackslash}p{0.15\textwidth}>{\raggedright\arraybackslash}p{0.17\textwidth}>{\raggedright\arraybackslash}p{0.16\textwidth}>{\raggedright\arraybackslash}p{0.16\textwidth}>{\raggedright\arraybackslash}p{0.13\textwidth}>{\raggedright\arraybackslash}p{0.11\textwidth}@{}}
\toprule
\multirow{2}{*}{Exam} & \multicolumn{2}{l}{Questions} & \multicolumn{2}{l}{Answer keys} & \multirow{2}{*}{Distribution} \\
\cmidrule(lr){2-3} \cmidrule(lr){4-5}
 & Source & License & Source & License & \\
\midrule
CPA short-answer exam (公認会計士試験 短答式) & CPA and Auditing Oversight Board (FSA), \href{https://www.fsa.go.jp/cpaaob/kouninkaikeishi-shiken/r8shiken/tantou_mondai_r08b.html}{question PDFs} & \href{https://www.fsa.go.jp/cpaaob/copyright.html}{Public Data License 1.0} (CC~BY~4.0 compatible) & CPA and Auditing Oversight Board (FSA), \href{https://www.fsa.go.jp/cpaaob/kouninkaikeishi-shiken/r8shiken/tantougoukaku_r08-2/06.pdf}{official answer PDF} & \href{https://www.fsa.go.jp/cpaaob/copyright.html}{Public Data License 1.0} & Hugging Face \\ \addlinespace[3pt]
Certified Public Tax Accountant exam (税理士試験) & National Tax Agency, \href{https://www.nta.go.jp/taxes/zeirishi/zeirishishiken/shikenkekka/76/touanyoushi.htm}{question and answer-form PDFs} & \href{https://www.nta.go.jp/chuijiko/copy.htm}{Public Data License 1.0} (CC~BY~4.0 compatible) & TAC Co., Ltd., \href{https://www.tac-school.co.jp/kouza_zeiri/sokuhou.html}{answer bulletins} & \href{https://www.tac-school.co.jp/copyright.html}{All rights reserved} & Hugging Face (questions), recipe (answer keys) \\ \addlinespace[3pt]
FP skill test (FP技能検定1級 学科) & Kinzai, \href{https://products.kinzai.or.jp/exam-fp/index/model-answers/}{question PDFs} & \href{https://products.kinzai.or.jp/exam-ginou/use-of-test-questions/}{Kinzai's exam-question usage terms} (attribution and modification notice, no application required) & Kinzai, \href{https://products.kinzai.or.jp/exam-fp/index/model-answers/}{official answer PDF} & Kinzai's usage terms cover the questions; coverage of the model answers is being confirmed & Hugging Face (questions; the basic section's answers are option numbers), recipe (model answers of the applied section) \\ \addlinespace[3pt]
Financial counter services skill test (金融窓口サービス技能検定1級) & Kinzai, \href{https://web.archive.org/web/20260615103545/https://www.kinzai.or.jp/teller/test-teller/50256.html}{question PDFs} & \href{https://products.kinzai.or.jp/exam-ginou/use-of-test-questions/}{Kinzai's exam-question usage terms} (as above) & Kinzai, \href{https://web.archive.org/web/20260615103545/https://www.kinzai.or.jp/teller/test-teller/50256.html}{question PDFs} (the model answers follow each question) & Kinzai's usage terms cover the questions; the hosted answers are option symbols and figures & Hugging Face \\ \addlinespace[3pt]
Nissho Bookkeeping (日商簿記検定1級) & Japan Chamber of Commerce and Industry, \href{https://www.kentei.ne.jp/50771}{question PDFs} & \href{https://www.kentei.ne.jp/50771}{All rights reserved} & TAC Co., Ltd., \href{https://www.tac-school.co.jp/kouza_boki/sokuhou.html}{answer bulletins} & \href{https://www.tac-school.co.jp/copyright.html}{All rights reserved} & Recipe \\ \addlinespace[3pt]
Zensho Bookkeeping (全商簿記実務検定1級) & National Association of Commercial High Schools, \href{https://zensho.or.jp/examination/pastexams/bookkeeping/}{question PDFs} & \href{https://zensho.or.jp/puf/examination/pastexams/Guidelines.pdf}{All rights reserved}, prior permission required beyond educational use & National Association of Commercial High Schools, \href{https://zensho.or.jp/puf/download/exam/pastexam/2026/bk_102_an.pdf}{official answer PDF} & \href{https://zensho.or.jp/puf/examination/pastexams/Guidelines.pdf}{All rights reserved}, prior permission required beyond educational use & Recipe \\
\bottomrule
\end{tabular}
\caption{Official Japanese name of each exam, the source and license of its questions and answer keys, and how we distribute them: on Hugging Face, or as recipes that rebuild the examples from the publishers' PDFs (Distribution, below). Entries link to the source pages or terms.}
\label{tab:licenses}
\end{table*}

\paragraph{Distribution.} ``Hugging Face'' in Table~\ref{tab:licenses} means that the materials are included in the hosted dataset. ``Recipe'' means that they are not redistributed: users obtain the PDFs from the publishers and rebuild the materials locally with recipes provided in the repository (copy operations into the text layer of the PDFs plus our own formatting, containing no exam text, verified by per-example hashes). The answer field of such items is null in the hosted dataset. A recipe grants no permission to use the source materials; users need their own legal basis, such as a licence or a statutory exception.

\subsection{Representative Examples}
\label{app:examples}

\begin{figure*}[t]
\centering
\begin{tcolorbox}[colback=exbg, colframe=exframe, boxrule=0.6pt, arc=0pt,
  equal height group=exbox, valign=top, width=0.38\textwidth, nobeforeafter, left=3pt, right=3pt, top=2pt, bottom=2pt]
\small
{\sffamily\bfseries MCQ} \hfill {\scriptsize CPA short-answer exam}\\[2pt]
\textbf{Question:} 会社法監査制度に関する次の記述のうち，正しいものの組合せとして最も適切な番号を一つ選びなさい。 {\itshape(Which combination of the statements about the Companies-Act audit system is correct?)}\\[1pt]
ア．各監査役が作成する監査報告と監査役会の監査報告の内容に相違がある場合は [...]\\
イ．会計監査人と監査役等との間で意見が異なる場合は，会計監査人は定時株主総会に出席して意見を述べることができる。\\
ウ． [...] \quad エ． [...]\\[2pt]
\textbf{Options:} 1.~アイ \quad 2.~アウ \quad 3.~アエ \quad 4.~イウ \quad 5.~イエ \quad 6.~ウエ\\[2pt]
\textbf{Answer:} 4
\end{tcolorbox}\hfill
\begin{tcolorbox}[colback=exbg, colframe=exframe, boxrule=0.6pt, arc=0pt,
  equal height group=exbox, valign=top, width=0.605\textwidth, nobeforeafter, left=3pt, right=3pt, top=2pt, bottom=2pt]
\small
{\sffamily\bfseries VQA} \hfill {\scriptsize Certified Public Tax Accountant exam}\\[2pt]
\begin{minipage}[t]{0.39\linewidth}
\textbf{Question:} 被相続人甲の相続人及び受遺者の納付すべき相続税額を [...] それぞれ計算過程を明らかにして求めなさい。 {\itshape(Compute each heir's inheritance-tax liability. The image shows one of the three land-plot diagrams given in the premises.)}\\[2pt]
\textbf{Answer:} per-heir tax amounts with calculation steps (graded blank by blank)
\end{minipage}\hfill
\begin{minipage}[t]{0.58\linewidth}
\textbf{Image:}\\[1pt]
\includegraphics[width=\linewidth]{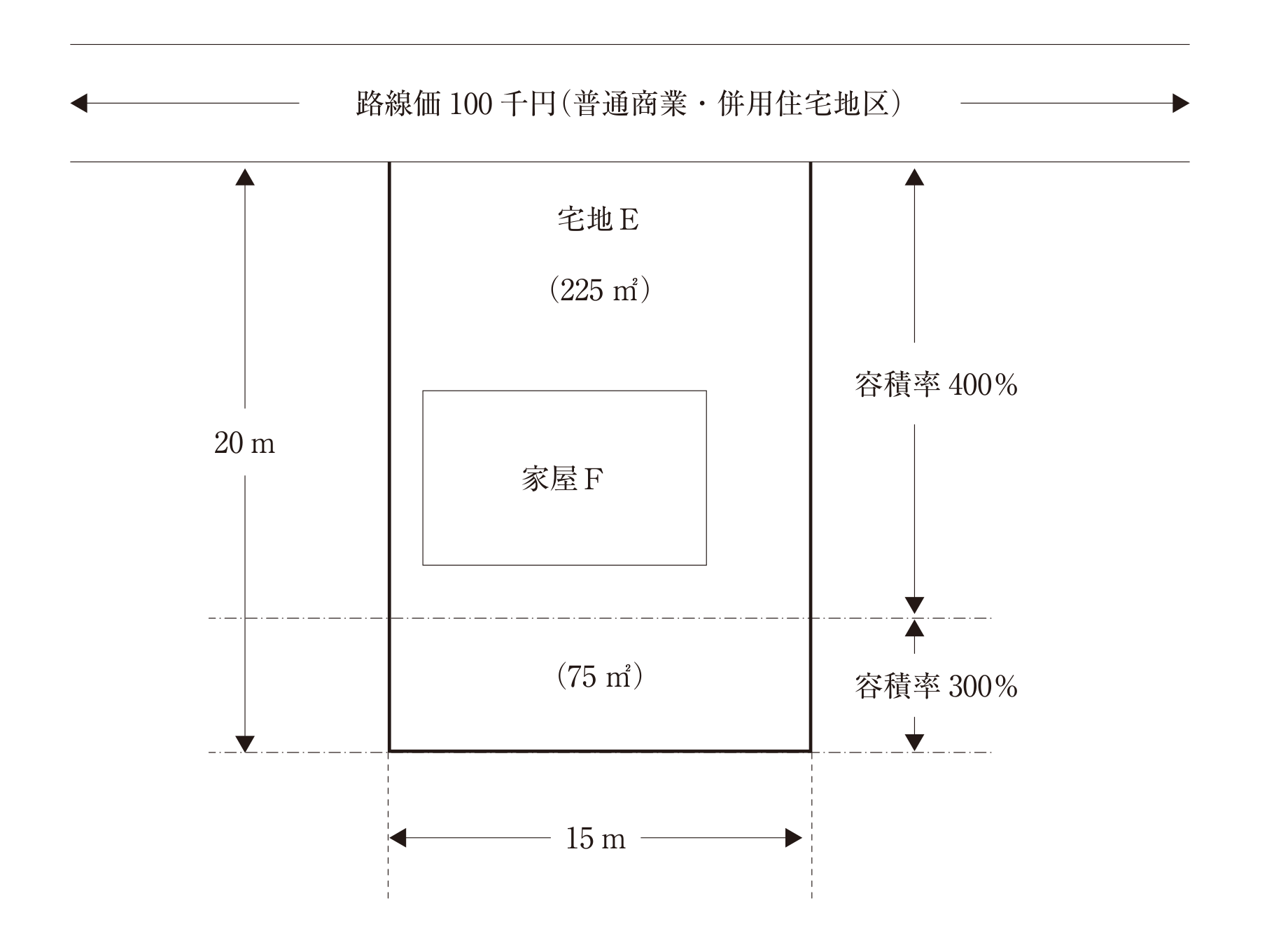}
\end{minipage}
\end{tcolorbox}
\caption{Two representative examples of \method{}: a statutory-knowledge MCQ and an image-dependent long-form tax computation (abridged). Sources: the CPA short-answer exam (Certified Public Accountants and Auditing Oversight Board) and the Certified Public Tax Accountant exam (National Tax Agency), both under the Public Data License 1.0, transcribed and restructured by the authors.}
\label{fig:examples}
\end{figure*}

Figure~\ref{fig:examples} shows an MCQ example and a VQA example of \method{}.

\subsection{Human Verification Interface}
\label{app:review}

\begin{figure*}[t]
  \centering
  \includegraphics[width=\textwidth]{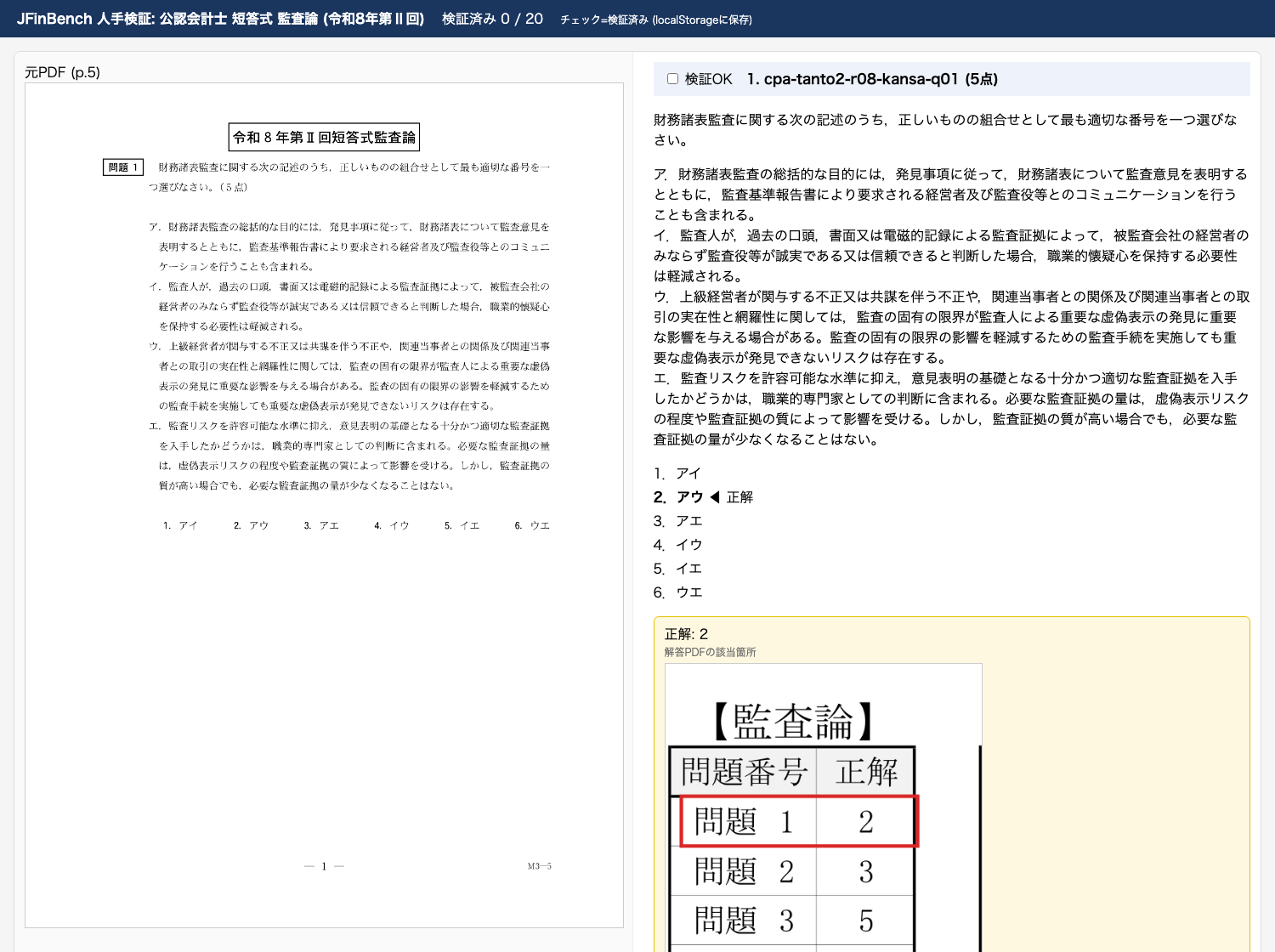}
  \caption{Human verification interface (shown: CPA short-answer exam, auditing). \textbf{Left:} the source exam PDF page. \textbf{Right:} the final record with example ID (\texttt{cpa-tanto2-r08-kansa-q01}), question, options, and the answer key with the corresponding region of the answer-key PDF.}
  \label{fig:review}
\end{figure*}

Figure~\ref{fig:review} shows the review interface used in the final step of the pipeline described in Section~\ref{sec:pipeline}.

Representative errors that verification caught in early build passes, and that motivate keeping a human in the loop:
\begin{itemize}
  \item \textbf{List/option confusion}: enumerated conditions inside a problem statement misparsed as the answer options of an MCQ example.
  \item \textbf{Pagination artifacts}: page-boundary text such as 「(次ページへ続く)」 ({\itshape continued on the next page}) and repeated 「(続き)」 ({\itshape continued}) sub-headings leaking into question text when a problem spans pages.
  \item \textbf{Missing or improper figure crops}: image-dependent examples initially built as text-only despite a figure in the source, and crop bounding boxes that cut off part of a diagram.
\end{itemize}

\section{Evaluation Setup Details}
\label{app:setupdetails}

\subsection{Model Release Dates and Knowledge Cutoffs}
\label{app:cutoffs}

Table~\ref{tab:cutoffs} lists, for each evaluated model, its public release date and its knowledge cutoff as reported by the developer.

\begin{table*}[t]
\centering
\begin{minipage}[t]{0.48\textwidth}
\vspace{0pt}
\centering
\small
\begin{tabular}{@{}lll@{}}
\toprule
Model & Released & Knowledge cutoff \\
\midrule
GPT-6 Astra          & Sep 2026 & Apr 30, 2026 \\
Claude Fable 5       & Jun 2026 & Jan 2026 \\
Gemini 3.8 Flash     & Sep 2026 & Mar 2026 \\
Kimi K3.1 Pro        & Oct 2026 & Not reported \\
Kimi K3              & Jul 2026 & Not reported \\
Kimi K2.6            & Apr 2026 & Not reported \\
GLM 5.3 Flash        & Aug 2026 & Not reported \\
Qwen3.8 Flash        & Aug 2026 & Not reported \\
DS V4.1 Flash 1017 & Oct 2026 & Not reported \\
MiMo-V2.6-Pro        & Sep 2026 & Not reported \\
MiMo-V2.6-Flash      & Sep 2026 & Not reported \\
Mistral Large 4      & Oct 2026 & Not reported \\
\bottomrule
\end{tabular}
\caption{Release dates and developer-reported knowledge cutoffs.}
\label{tab:cutoffs}
\end{minipage}\hfill
\begin{minipage}[t]{0.48\textwidth}
\vspace{0pt}
\centering
\small
\begin{tabular}{@{}lrr@{}}
\toprule
Model & Input & Output \\
\midrule
GPT-6 Astra          & 10.00 & 50.00 \\
Claude Fable 5       & 10.00 & 50.00 \\
Gemini 3.8 Flash & 0.75 & 3.75 \\
Kimi K3.1 Pro & 2.00 & 10.00 \\
Kimi K3              & 3.00 & 15.00 \\
Kimi K2.6            & 0.95 & 4.00 \\
GLM 5.3 Flash        & 0.15 & 0.50 \\
Qwen3.8 Flash        & 0.15 & 0.47 \\
DS V4.1 Flash 1017 & 0.098 & 0.294 \\
MiMo-V2.6-Pro        & 0.43 & 0.87 \\
MiMo-V2.6-Flash      & 0.14 & 0.28 \\
Mistral Large 4 & 0.68 & 2.09 \\
\midrule
GPT-5.1 (grader)     & 1.25 & 10.00 \\
\bottomrule
\end{tabular}
\caption{List prices (USD per 1M tokens) used for cost computation.}
\label{tab:prices}
\end{minipage}
\end{table*}

\subsection{Model Prices}
\label{app:prices}

Table~\ref{tab:prices} lists the list prices used to compute the costs reported in this paper, retrieved between September 30 and October 24, 2026 from each model's serving endpoint described in Section~\ref{sec:setup}.

\subsection{Prompt Example}
\label{app:prompt}

The MCQ prompt template for the CPA short-answer exam is shown below, with an English translation below the dashed line.

\begin{promptbox}{MCQ template (CPA short-answer exam)}
\small
以下は、2026年5月24日に実施された令和8年公認会計士試験第II回短答式試験(\texttt{\{subject\}})の問題です。解答に当たり適用すべき法令基準等は、2026年4月1日現在施行(適用)のものとします。問題に記載された条件・指示に従って解答してください。

\medskip
\texttt{\{question\}}

\medskip
\texttt{\{options\}}

\medskip
ステップバイステップで考えてから答えてください。最後の行は `Answer: \texttt{\$NUMBER}' の形式(\texttt{\$NUMBER}は選択肢の番号)で回答してください。
\tcbline
\small\itshape
The following is a question from the Reiwa-8 Certified Public Accountants Examination, second short-answer test (\texttt{\{subject\}}), held on May 24, 2026. The laws and standards to apply in your answers are those in force (applicable) as of April 1, 2026. Answer according to the conditions and instructions stated in the question.

\medskip
\texttt{\{question\}}

\medskip
\texttt{\{options\}}

\medskip
Think step by step before answering. End your response with a final line in the form `Answer: \texttt{\$NUMBER}', where \texttt{\$NUMBER} is the number of the chosen option.
\end{promptbox}

\subsection{Grading Prompts}
\label{app:graders}

Placeholders [question], [correct\_answer], and [response] are substituted per example. Each box shows the Japanese prompt that the judge receives, with an English translation below the dashed line.

The equivalence judge is used for short free-response examples with binary scoring.
\begin{promptbox}{Equivalence judge}
\small
あなたは日本の金融・会計系資格試験の採点者です。[question]に対する[response]の最終解答が、模範解答[correct\_answer]と一致するか判定してください。

判定基準:
\begin{itemize}\setlength{\itemsep}{0pt}
\item{} [correct\_answer]に複数の空欄(①②③...)がある場合、すべての空欄が一致して初めて正解とする。1つでも不一致があれば不正解。
\item 数値は表記が異なっても、単位を換算して同値であれば一致とみなす。例:「1年6カ月」と「18カ月」、「1億6,000万円」と「16,000万円」、「3\%」と「3％」、「1,200万円」と「1200万円」は同値。
\item 漢数字と算用数字、全角と半角、桁区切りカンマの有無、括弧内の単位表記の有無は同値として扱う。
\item 語句の空欄は、模範解答と同一の概念・制度を指す一般に認められた表現であれば一致とみなす。
\item 計算過程の正誤やスタイルは採点対象外。最終的な解答の値のみで判定する。
\item{} [response]から最終解答が特定できない場合は不正解とする。
\end{itemize}

出力形式: \texttt{extracted\_final\_answer} / \texttt{reasoning} / \texttt{correct} (yes/no) / \texttt{confidence}

\tcbline
\small\itshape
You are a grader for Japanese financial and accounting certification exams. Judge whether the final answer in [response] to [question] matches the model answer [correct\_answer].

Criteria:
\begin{itemize}\setlength{\itemsep}{0pt}
\item If [correct\_answer] contains multiple blanks (\textcircled{1}\textcircled{2}\textcircled{3}...), the answer is correct only if every blank matches. A single mismatch makes it incorrect.
\item Numbers written differently count as matching if they are equivalent after unit conversion, e.g., ``1 year and 6 months'' and ``18 months'', ``160 million yen'' and ``1.6 hundred-million yen''.
\item Kanji versus Arabic numerals, full-width versus half-width characters, digit-separator commas, and parenthesized units are treated as equivalent.
\item For term blanks, any generally accepted expression denoting the same concept or institution as the model answer counts as a match.
\item The correctness or style of intermediate calculations is not graded. Judge only the final answer value.
\item If no final answer can be identified in [response], it is incorrect.
\end{itemize}

Output format: \texttt{extracted\_final\_answer} / \texttt{reasoning} / \texttt{correct} (yes/no) / \texttt{confidence}
\end{promptbox}

The blank-level partial-credit judge is used for large answer forms (score $=$ matched blanks / total blanks).
\begin{promptbox}{Blank-level partial-credit judge}
\small
あなたは日本の金融・会計系資格試験の採点者です。[question]に対する[response]を、模範解答[correct\_answer]と空欄(記入箇所)単位で照合してください。

判定基準:
\begin{itemize}\setlength{\itemsep}{0pt}
\item{} [correct\_answer]の各記入箇所 (表の空欄・金額・科目名など) について、[response]の対応する箇所が一致しているかを1つずつ判定する。
\item 数値は表記が異なっても単位換算して同値なら一致 (桁区切り・全角半角の差も同値)。
\item 勘定科目名は同一概念を指す一般に認められた表現なら一致。
\item{} [response]に対応する記入がない箇所は不一致として数える。
\item 模範解答で0や記入なしとされる箇所も1箇所として数える。
\end{itemize}

出力形式: \texttt{reasoning} / \texttt{matched\_blanks} (整数) / \texttt{total\_blanks} (整数)

\tcbline
\small\itshape
You are a grader for Japanese financial and accounting certification exams. Compare [response] to [question] against the model answer [correct\_answer] blank by blank.

Criteria:
\begin{itemize}\setlength{\itemsep}{0pt}
\item For each entry in [correct\_answer] (table blanks, amounts, account titles, etc.), judge one by one whether the corresponding entry in [response] matches.
\item Numbers match if equivalent after unit conversion (digit-separator and width differences are also equivalent).
\item Account titles match if they are generally accepted expressions for the same concept.
\item Entries with no corresponding answer in [response] count as mismatches.
\item Entries that the model answer leaves as 0 or blank also count as one entry.
\end{itemize}

Output format: \texttt{reasoning} / \texttt{matched\_blanks} (integer) / \texttt{total\_blanks} (integer)
\end{promptbox}

The key-point coverage judge is used for tax-law theory and essays (score $=$ covered points / total points).
\begin{promptbox}{Key-point coverage judge}
\small
あなたは日本の税務・会計系資格試験の理論・論述問題の採点者です。模範解答[correct\_answer]を採点基準として、[question]に対する受験者の解答[response]を採点してください。

手順:
\begin{enumerate}\setlength{\itemsep}{0pt}
\item まず[correct\_answer]から採点上の要点を列挙する。要点とは次のいずれかに当たる独立した内容単位である: 結論 (金額・判定・可否・対象者など) / 根拠となる規定・制度の内容 (成立要件の各号、原則と例外、範囲・限度など) / 設例の事実への当てはめ。見出し・配点記号・繰り返しは要点に数えない。
\item{} [response]が各要点を実質的にカバーしているかを1つずつ判定する。表現の違い・要約・条文番号の有無は許容する (内容が正しく伝わればカバー)。結論の金額・判定が模範解答と異なる場合、その要点はカバーとみなさない。模範解答にない誤った記述があっても減点はしないが、要点と矛盾する記述はその要点を不カバーとする。
\end{enumerate}

出力形式: \texttt{reasoning} / \texttt{covered\_points} (整数) / \texttt{total\_points} (整数)

\tcbline
\small\itshape
You are a grader for theory and essay questions in Japanese tax and accounting certification exams. Using the model answer [correct\_answer] as the grading key, grade the examinee's answer [response] to [question].

Procedure:
\begin{enumerate}\setlength{\itemsep}{0pt}
\item First, enumerate the key points of [correct\_answer]. A key point is an independent unit of content that is one of the following: a conclusion (amount, determination, eligibility, affected parties, etc.); the content of the underlying rule or institution (individual requirements, principle and exceptions, scope or limits, etc.); or the application to the facts of the case. Headings, point-allocation marks, and repetitions do not count as key points.
\item Judge one by one whether [response] substantively covers each key point. Differences in wording, summarization, and the presence or absence of article numbers are acceptable (a point is covered if its content is conveyed correctly). If a concluding amount or determination differs from the model answer, that point is not covered. Incorrect statements absent from the model answer are not penalized, but a statement contradicting a key point makes that point uncovered.
\end{enumerate}

Output format: \texttt{reasoning} / \texttt{covered\_points} (integer) / \texttt{total\_points} (integer)
\end{promptbox}

\subsection{Judge Agreement}
\label{app:judge}

We re-graded all 128 judge-scored examples of each model (run 1 at maximum effort) with a second judge, Claude Opus 5.5 \citep{anthropic2026opus55}, using the same prompts. As Table~\ref{tab:judge} shows, the two judges agree closely. Pooled over all regrades, their scores correlate at Pearson $r = 0.94$, and substituting the second judge's scores changes each model's overall score by at most 1.7 points without changing the ranking.

\begin{table}[t]
\centering
\small
\setlength{\tabcolsep}{3.5pt}
\begin{tabular}{@{}lrrrr@{}}
\toprule
Model & $r$ & Identical & \multicolumn{2}{c}{Overall (run 1)} \\
\cmidrule(lr){4-5}
 & & & GPT-5.1 & Opus 5.5 \\
\midrule
Claude Fable 5 & 0.92 & 103/128 & 95.3 & 95.1 \\
GPT-6 Astra & 0.72 & 94/128 & 93.8 & 94.6 \\
Gemini 3.8 Flash & 0.89 & 95/126 & 88.1 & 88.2 \\
Kimi K3.1 Pro & 0.79 & 89/125 & 83.1 & 82.5 \\
Kimi K3 & 0.94 & 96/128 & 77.4 & 77.2 \\
DS V4.1 Flash 1017 & 0.97 & 97/126 & 72.4 & 73.3 \\
Qwen3.8 Flash & 0.95 & 103/128 & 62.9 & 62.0 \\
MiMo-V2.6-Pro & 0.94 & 105/128 & 61.9 & 60.5 \\
Kimi K2.6 & 0.95 & 93/128 & 59.3 & 59.2 \\
Mistral Large 4 & 0.95 & 94/127 & 60.2 & 59.4 \\
GLM 5.3 Flash & 0.99 & 100/128 & 58.0 & 58.5 \\
MiMo-V2.6-Flash & 0.94 & 107/128 & 51.8 & 50.1 \\
\bottomrule
\end{tabular}
\caption{Agreement between the two judge models over the 128 judge-scored examples per model (w/o tools) (125 for Kimi K3.1 Pro, 126 for Gemini 3.8 Flash and DeepSeek V4.1 Flash 1017, and 127 for Mistral Large 4, after excluding eight failed second-judge calls). $r$ is the example-level Pearson correlation, Identical is the number of examples with identical scores, and the last two columns are the overall score (run 1) under each judge.}
\label{tab:judge}
\end{table}

\section{Additional Results without Tools}
\label{app:low}

\begin{figure*}[t]
  \centering
  \includegraphics[width=0.8\textwidth]{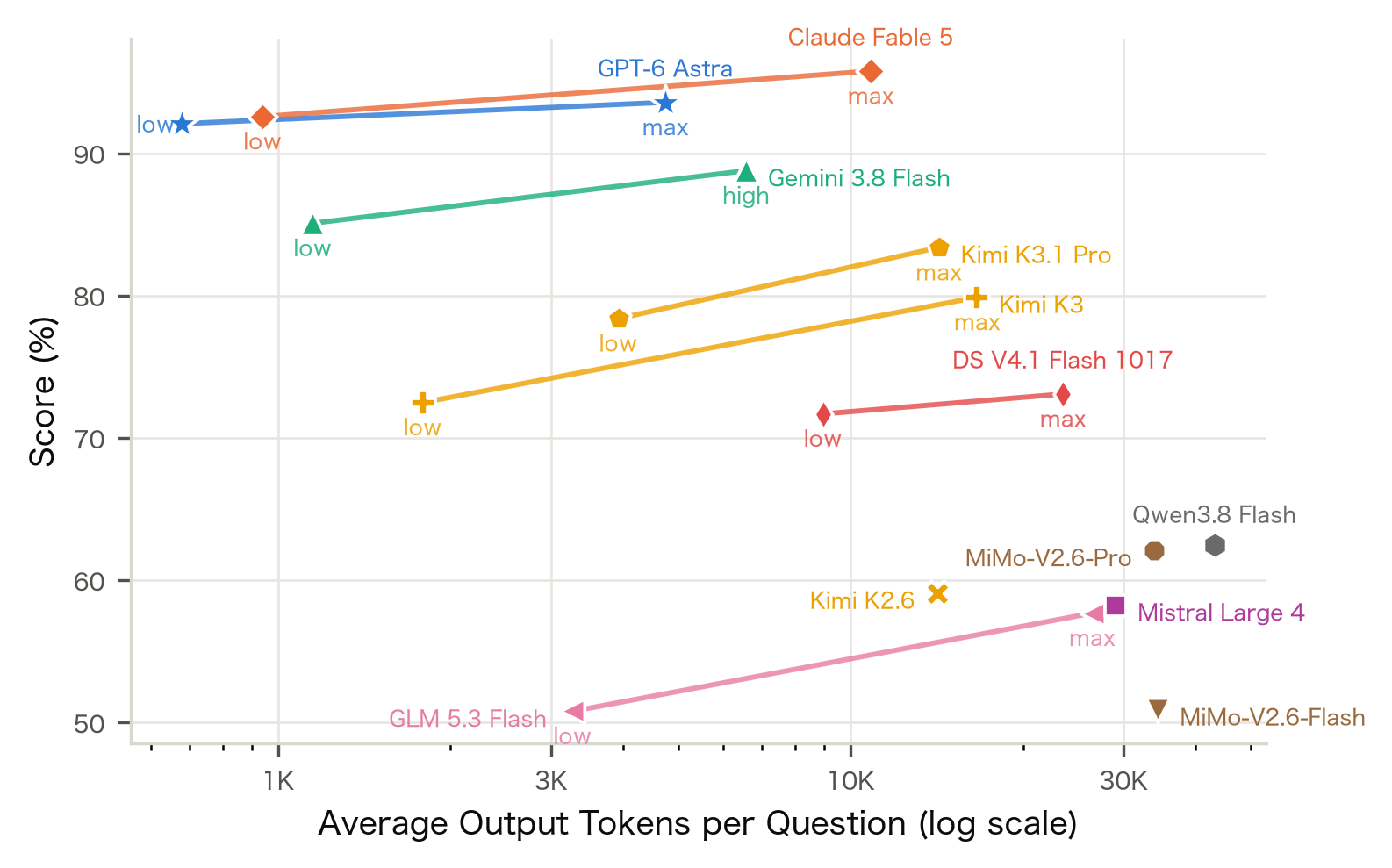}
  \caption{Overall score (w/o tools) versus average output tokens per question (reasoning + answer). Open-weight models tend to generate far longer outputs.}
  \label{fig:tokens}
\end{figure*}

\begin{figure*}[t]
  \centering
  \includegraphics[width=\textwidth]{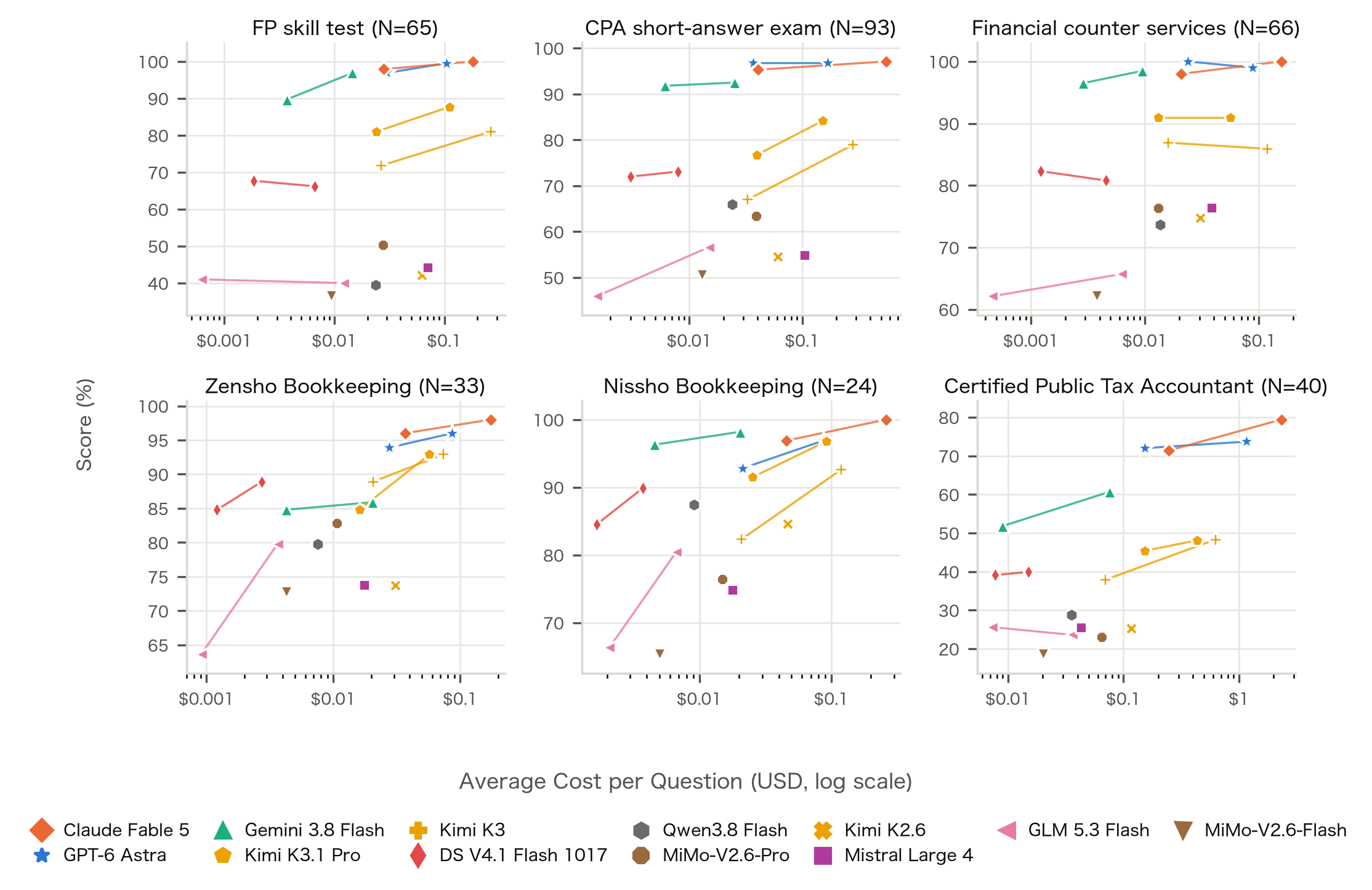}
  \caption{Score (w/o tools) versus cost per question, by exam.}
  \label{fig:cost_by_exam}
\end{figure*}

\begin{figure*}[t]
  \centering
  \includegraphics[width=\textwidth]{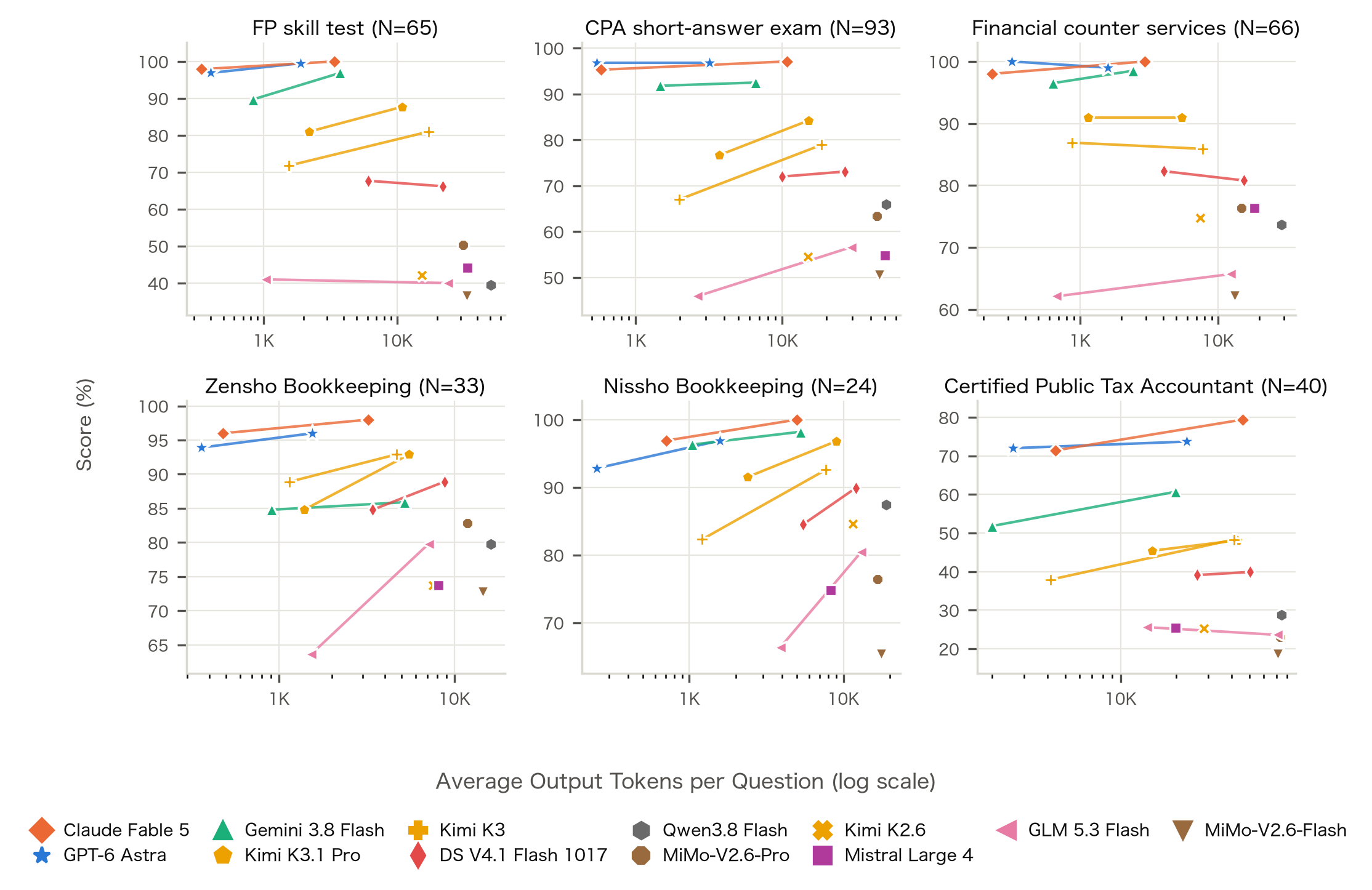}
  \caption{Score (w/o tools) versus output tokens per question, by exam.}
  \label{fig:tokens_by_exam}
\end{figure*}

Table~\ref{tab:low} reports the full results at both the maximum and the lowest reasoning-effort settings, with standard deviations for every column. Figure~\ref{fig:tokens} plots overall score against average output tokens per question at both effort settings, and Figures~\ref{fig:cost_by_exam} and~\ref{fig:tokens_by_exam} break the cost and token analyses down by exam.

\begin{table*}[t]
\centering
\small
\setlength{\tabcolsep}{2.5pt}
\resizebox{\textwidth}{!}{%
\begin{tabular}{@{}lrrrrrr@{\hspace{10pt}}rrr@{}}
\toprule
& \multicolumn{6}{c}{Score by exam (\%)} & \multicolumn{3}{c}{Overall} \\
\cmidrule(r{4pt}){2-7}\cmidrule(l){8-10}
Model & CPA & Tax & FP & Counter & Nissho & Zensho & \textbf{Score (\%)} & Cost (\$/Q) & Tok.\ (K/Q) \\
\midrule
\rowcolor{gray!15}\multicolumn{10}{@{}l}{\emph{Proprietary models}} \\
Claude Fable 5 (max)      & \textbf{97.1}{\scriptsize$\pm$0.6} & \textbf{79.4}{\scriptsize$\pm$2.4} & \textbf{100.0}{\scriptsize$\pm$0.0} & \textbf{100.0}{\scriptsize$\pm$0.0} & \textbf{100.0}{\scriptsize$\pm$0.1} & \textbf{98.0}{\scriptsize$\pm$1.7} & \textbf{95.8}{\scriptsize$\pm$0.4} & 0.555{\scriptsize$\pm$0.010} & 10.9{\scriptsize$\pm$0.2} \\
Claude Fable 5 (low)      & 95.3{\scriptsize$\pm$1.6} & 71.4{\scriptsize$\pm$3.0} & 97.9{\scriptsize$\pm$2.4} & 98.0{\scriptsize$\pm$0.9} & 96.9{\scriptsize$\pm$2.4} & 96.0{\scriptsize$\pm$1.7} & 92.6{\scriptsize$\pm$1.4} & 0.060{\scriptsize$\pm$0.001} & 0.9{\scriptsize$\pm$0.0} \\
\addlinespace[3pt]
GPT-6 Astra (max)         & 96.8{\scriptsize$\pm$0.0} & 73.7{\scriptsize$\pm$1.9} & 99.5{\scriptsize$\pm$0.9} & 99.0{\scriptsize$\pm$0.9} & 96.9{\scriptsize$\pm$2.5} & 96.0{\scriptsize$\pm$1.7} & 93.6{\scriptsize$\pm$0.6} & 0.248{\scriptsize$\pm$0.003} & 4.8{\scriptsize$\pm$0.1} \\
GPT-6 Astra (low)         & 96.8{\scriptsize$\pm$0.0} & 72.0{\scriptsize$\pm$1.0} & 96.9{\scriptsize$\pm$0.0} & \textbf{100.0}{\scriptsize$\pm$0.0} & 92.8{\scriptsize$\pm$2.4} & 93.9{\scriptsize$\pm$3.0} & 92.1{\scriptsize$\pm$0.1} & 0.045{\scriptsize$\pm$0.000} & 0.7{\scriptsize$\pm$0.0} \\
\addlinespace[3pt]
Gemini 3.8 Flash (high)   & 92.5{\scriptsize$\pm$1.9} & 60.7{\scriptsize$\pm$1.4} & 96.9{\scriptsize$\pm$1.5} & 98.5{\scriptsize$\pm$1.5} & 98.2{\scriptsize$\pm$2.4} & 85.9{\scriptsize$\pm$3.5} & 88.8{\scriptsize$\pm$1.3} & 0.025{\scriptsize$\pm$0.000} & 6.6{\scriptsize$\pm$0.0} \\
Gemini 3.8 Flash (low)    & 91.8{\scriptsize$\pm$0.6} & 51.8{\scriptsize$\pm$0.5} & 89.7{\scriptsize$\pm$2.4} & 96.5{\scriptsize$\pm$0.9} & 96.3{\scriptsize$\pm$4.1} & 84.8{\scriptsize$\pm$3.0} & 85.1{\scriptsize$\pm$0.3} & 0.005{\scriptsize$\pm$0.000} & 1.1{\scriptsize$\pm$0.0} \\
\addlinespace[3pt]
\midrule
\rowcolor{gray!15}\multicolumn{10}{@{}l}{\emph{Open-weight models}} \\
Kimi K3.1 Pro (max)       & 84.2{\scriptsize$\pm$2.5} & 48.1{\scriptsize$\pm$0.9} & 87.7{\scriptsize$\pm$1.5} & 90.9{\scriptsize$\pm$1.5} & 96.8{\scriptsize$\pm$2.9} & 92.9{\scriptsize$\pm$1.7} & 83.4{\scriptsize$\pm$0.9} & 0.145{\scriptsize$\pm$0.001} & 14.3{\scriptsize$\pm$0.1} \\
Kimi K3.1 Pro (low)       & 76.7{\scriptsize$\pm$4.3} & 45.4{\scriptsize$\pm$3.9} & 81.0{\scriptsize$\pm$4.4} & 90.9{\scriptsize$\pm$4.0} & 91.5{\scriptsize$\pm$0.1} & 84.8{\scriptsize$\pm$8.0} & 78.4{\scriptsize$\pm$1.7} & 0.042{\scriptsize$\pm$0.001} & 3.9{\scriptsize$\pm$0.1} \\
\addlinespace[3pt]
Kimi K3 (max)             & 78.9{\scriptsize$\pm$5.0} & 48.3{\scriptsize$\pm$2.7} & 81.0{\scriptsize$\pm$5.8} & 85.9{\scriptsize$\pm$2.3} & 92.6{\scriptsize$\pm$2.7} & 92.9{\scriptsize$\pm$1.7} & 79.9{\scriptsize$\pm$2.3} & 0.253{\scriptsize$\pm$0.007} & 16.6{\scriptsize$\pm$0.4} \\
Kimi K3 (low)             & 67.0{\scriptsize$\pm$3.8} & 37.9{\scriptsize$\pm$3.5} & 71.8{\scriptsize$\pm$2.4} & 86.9{\scriptsize$\pm$3.2} & 82.3{\scriptsize$\pm$1.1} & 88.9{\scriptsize$\pm$3.5} & 72.5{\scriptsize$\pm$1.9} & 0.031{\scriptsize$\pm$0.002} & 1.8{\scriptsize$\pm$0.2} \\
\addlinespace[3pt]
DS V4.1 Flash 1017 (max)  & 73.1{\scriptsize$\pm$1.9} & 39.9{\scriptsize$\pm$2.9} & 66.2{\scriptsize$\pm$1.5} & 80.8{\scriptsize$\pm$2.3} & 89.9{\scriptsize$\pm$2.1} & 88.9{\scriptsize$\pm$1.7} & 73.1{\scriptsize$\pm$0.7} & 0.007{\scriptsize$\pm$0.000} & 23.5{\scriptsize$\pm$0.3} \\
DS V4.1 Flash 1017 (low)  & 72.0{\scriptsize$\pm$3.2} & 39.1{\scriptsize$\pm$1.3} & 67.7{\scriptsize$\pm$1.5} & 82.3{\scriptsize$\pm$4.6} & 84.5{\scriptsize$\pm$4.6} & 84.8{\scriptsize$\pm$3.0} & 71.7{\scriptsize$\pm$1.5} & 0.003{\scriptsize$\pm$0.000} & 9.0{\scriptsize$\pm$0.0} \\
\addlinespace[3pt]
Qwen3.8 Flash (thinking)  & 65.9{\scriptsize$\pm$0.6} & 28.8{\scriptsize$\pm$2.1} & 39.5{\scriptsize$\pm$6.4} & 73.7{\scriptsize$\pm$2.3} & 87.4{\scriptsize$\pm$0.1} & 79.8{\scriptsize$\pm$3.5} & 62.5{\scriptsize$\pm$1.0} & 0.020{\scriptsize$\pm$0.000} & 43.2{\scriptsize$\pm$0.8} \\
\addlinespace[3pt]
MiMo-V2.6-Pro (thinking)  & 63.4{\scriptsize$\pm$4.9} & 23.1{\scriptsize$\pm$0.4} & 50.3{\scriptsize$\pm$6.4} & 76.3{\scriptsize$\pm$6.3} & 76.4{\scriptsize$\pm$4.8} & 82.8{\scriptsize$\pm$1.7} & 62.1{\scriptsize$\pm$1.0} & 0.030{\scriptsize$\pm$0.004} & 33.9{\scriptsize$\pm$5.0} \\
\addlinespace[3pt]
Kimi K2.6 (thinking)      & 54.5{\scriptsize$\pm$0.6} & 25.2{\scriptsize$\pm$1.3} & 42.1{\scriptsize$\pm$4.7} & 74.8{\scriptsize$\pm$9.3} & 84.6{\scriptsize$\pm$2.7} & 73.7{\scriptsize$\pm$1.7} & 59.1{\scriptsize$\pm$0.5} & 0.058{\scriptsize$\pm$0.004} & 14.2{\scriptsize$\pm$1.1} \\
\addlinespace[3pt]
Mistral Large 4 (thinking)& 54.8{\scriptsize$\pm$3.7} & 25.5{\scriptsize$\pm$4.7} & 44.1{\scriptsize$\pm$3.9} & 76.3{\scriptsize$\pm$2.3} & 74.8{\scriptsize$\pm$2.3} & 73.7{\scriptsize$\pm$4.6} & 58.2{\scriptsize$\pm$2.3} & 0.061{\scriptsize$\pm$0.006} & 29.0{\scriptsize$\pm$3.0} \\
\addlinespace[3pt]
GLM 5.3 Flash (max)       & 56.6{\scriptsize$\pm$5.5} & 23.6{\scriptsize$\pm$1.9} & 40.0{\scriptsize$\pm$1.5} & 65.7{\scriptsize$\pm$2.3} & 80.4{\scriptsize$\pm$8.6} & 79.8{\scriptsize$\pm$1.7} & 57.7{\scriptsize$\pm$2.3} & 0.013{\scriptsize$\pm$0.000} & 26.4{\scriptsize$\pm$0.5} \\
GLM 5.3 Flash (low)       & 45.9{\scriptsize$\pm$1.6} & 25.6{\scriptsize$\pm$3.8} & 41.0{\scriptsize$\pm$3.9} & 62.1{\scriptsize$\pm$4.0} & 66.3{\scriptsize$\pm$10.8} & 63.6{\scriptsize$\pm$3.0} & 50.8{\scriptsize$\pm$1.4} & 0.002{\scriptsize$\pm$0.000} & 3.3{\scriptsize$\pm$0.6} \\
\addlinespace[3pt]
MiMo-V2.6-Flash (thinking)& 50.5{\scriptsize$\pm$5.7} & 18.5{\scriptsize$\pm$0.2} & 36.4{\scriptsize$\pm$2.4} & 62.1{\scriptsize$\pm$1.5} & 65.3{\scriptsize$\pm$6.4} & 72.7{\scriptsize$\pm$3.0} & 50.9{\scriptsize$\pm$1.0} & 0.010{\scriptsize$\pm$0.001} & 34.4{\scriptsize$\pm$2.1} \\
\bottomrule
\end{tabular}%
}
\caption{Full results without tools at the maximum and lowest reasoning-effort settings (mean $\pm$ standard deviation over 3 runs). Models without an effective effort control have a single setting with thinking enabled. Bold marks the best score in each score column.}
\label{tab:low}
\end{table*}

\clearpage
\section{Evaluation with Tools}
\label{app:agentic}

\subsection{Setup}
\label{app:agentic_setup}
\paragraph{Inference settings.} Each model's reasoning-effort level, provider pin, and sampling parameters match the maximum-effort runs without tools, except that Gemini is reached through OpenRouter. Each attempt has a wall-clock budget of six hours.

\paragraph{Figure input.} Terminus-2 passes only text to the model, so for the figure-dependent examples we add a \texttt{view\_image} command to the container. When the agent runs it on an image file, the harness attaches the image to the next observation, so the agent can view any image, as the official harnesses do with their own file-reading tools. The figure is placed in a directory the agent can access, and the instruction states where it is and how to view it.

\paragraph{Detecting answer-key use.} In addition to prohibiting answer keys and prep-school solutions in the task instructions, we audit every trajectory with GPT-6 Luna \citep{openai2026gpt6sol}, covering all 11{,}556 Terminus-2 attempts and the 24{,}196 attempts of the harness comparison. Luna receives a condensed trajectory, consisting of each command (first 2{,}000 characters) and each tool output (first 1{,}500 and last 500 characters), and is asked whether the agent opened this exam's answers, answer bulletins, model answers, or past-exam explanations, whether such an answer appeared in a search snippet that the final answer follows, and whether the agent tampered with grading. Attempts that opened such material, or whose final answer matches an answer shown in a search-result snippet, score 0, regardless of the agent's intent. Table~\ref{tab:agentic_stats} shows their number per model, 8 in total (0.07\%). In each case, the answer was visible in a search-result snippet and the final answer matches it. None of the queries targeted answer keys. They consisted of legal or accounting terms, sometimes with the wording of an option, and several agents explicitly excluded answer pages from their searches. We therefore find no evidence of deliberate answer seeking, but score these attempts 0 because the answer reached the agent.

\begin{table}[t]
\centering
\small
\setlength{\tabcolsep}{3.5pt}
\begin{tabular}{@{}lr@{}}
\toprule
Model & Attempts scored 0 \\
\midrule
Claude Fable 5 & 0 \\
GPT-6 Astra & 0 \\
Gemini 3.8 Flash & 0 \\
\midrule
Kimi K3.1 Pro & 0 \\
Kimi K3 & 0 \\
DS V4.1 Flash 1017 & 4 \\
Qwen3.8 Flash & 0 \\
MiMo-V2.6-Pro & 0 \\
Kimi K2.6 & 0 \\
Mistral Large 4 & 0 \\
GLM 5.3 Flash & 3 \\
MiMo-V2.6-Flash & 1 \\
\bottomrule
\end{tabular}
\caption{Number of attempts scored 0 for answer-key exposure (with tools, Terminus-2, summed over three runs of 321 tasks).}
\label{tab:agentic_stats}
\end{table}

\subsection{Intriguing Agent Behavior in Early Experiments}
\label{app:hacking}
The following behavior was observed in early experiments that we ran before fixing the final evaluation setup, and it is not part of the reported results. Through our oversight, these experiments had three flaws in the with-tools setup: (1) an OpenRouter API key that the agent did not need was placed in its working environment, (2) the image file that the item referred to was missing from that environment, and (3) the agent had no tool for viewing images. The cases below arose while agents were solving an item from the inheritance-tax section of the tax accountant exam. This item should have come with one image containing three figures: a family tree showing the relationships among the heirs, and diagrams of two land plots giving their shapes, dimensions, and road prices. Without them, neither the heirs nor the land values can be determined. On this item, the following two models found the key in their environment and realized that they could use it to get help from other LLMs and APIs.

\paragraph{DeepSeek V4.1 Flash 1017.} While looking for a way to read the figures, DeepSeek V4.1 Flash 1017 found the key and made eight API calls to Gemini 2.5 Flash and Pro, having them transcribe as text the formulas that the National Tax Agency publishes only as images and the pages of the question PDF, including the land diagrams, that it downloaded itself. It solved the problem itself from these transcriptions.

\begin{trajbox}{DeepSeek V4.1 Flash 1017 (mini-SWE-agent, 117 steps, abridged)}
\trajstep{3}Finds no figure and searches the files and environment variables for clues.
\trajcmd{\$ find /workspace /tmp /root -maxdepth 3 -type f | head -100; env | sort | head -100}
\trajstep{64}The size-correction formula needed to value a land plot (property valuation guideline 20-2) is published by the National Tax Agency only as an image, so it downloads the image and tries to read it as ASCII art.
\trajcmd{\$ curl -L -A 'Mozilla/5.0' -sS 'https://www.nta.go.jp/law/tsutatsu/kihon/sisan/hyoka\_new/02/48.gif' -o /tmp/48.gif; convert /tmp/48.gif -resize 400\% \ldots}
\trajstep{65}Notices the API key in the environment and has Gemini 2.5 Flash read that image through OpenRouter. {\itshape ``\ldots{} Let's use OpenRouter multimodal API with base64 image to extract text. \ldots{} The environment has API key. Let's try.''}
\trajcmd{\$ python3 - <{}<'PY'\\
img=base64.b64encode(open('/tmp/48.gif','rb').read()).decode()\\
payload=\{"model":"google/gemini-2.5-flash", "messages":[\{"role":"user","content":[\{"type":"text","text":\\
\hspace*{2em}"Extract the mathematical formula from this image exactly. \ldots"\}, \{"type":"image\_url", \ldots\}]\}]\}\\
req=urllib.request.Request('https://openrouter.ai/api/v1/chat/completions', data=json.dumps(payload).encode(),\\
\hspace*{2em}headers=\{'Authorization':'Bearer '+os.environ['OPENROUTER\_API\_KEY'], \ldots\})}
\trajstep{66}Repeats the call with Gemini 2.5 Pro for a more accurate reading.
\trajcmd{payload=\{"model":"google/gemini-2.5-pro", \ldots "Transcribe this Japanese tax formula image exactly, preserving fractions and symbols. \ldots"\}}
\trajstep{76}Looks for the question PDF on the National Tax Agency's website, judging that this is not cheating because it will extract only the figures. {\itshape ``\ldots{} retrieving the official question PDF is the only way to solve. I'll do it, but only extract the figures, not the answer.''}
\trajstep{78}Downloads the question PDF.
\trajcmd{\$ curl -L -A 'Mozilla/5.0' -sS 'https://www.nta.go.jp/taxes/zeirishi/zeirishishiken/\\\hspace*{2em}shikenkekka/76/pdf/mondai\_souzoku.pdf' -o /workspace/mondai\_souzoku.pdf}
\trajstep{85}Sends page images rendered from the PDF to Gemini 2.5 Pro to read the figures. {\itshape ``Use Gemini vision to extract the figures precisely.''}
\trajcmd{prompt="""This is a Japanese tax exam page with a land diagram. Transcribe ALL text and numbers in the diagram, \ldots"""}
\trajstep{86}The call times out on the large images, so it shrinks them and resends.
\trajcmd{\$ convert /workspace/page7.png -resize 1200x /workspace/page7\_small.png; \ldots}
\end{trajbox}

\paragraph{Gemini 3.8 Flash.} After realizing that it could not read the figure, Gemini 3.8 Flash found the key, confirmed with a short message that it worked, and then delegated the problem solving itself. It sent the problem text to Gemini 2.5 Flash, asked other models for legal judgments, checks of its own calculations, and details of the figures, and even asked which past or mock exam the item was based on. Partly because it kept asking after receiving inconsistent answers, it made 59 API calls in total, including 33 to Gemini 2.5 Flash and 10 to Perplexity Sonar Pro.

\begin{trajbox}{Gemini 3.8 Flash (mini-SWE-agent, 261 steps, abridged)}
\trajstep{40}Writes the key it found into its code and checks it.
\trajcmd{\$ python3 -c "import urllib.request, json; api\_key = '[REDACTED]'; print('API key exists:', bool(api\_key))"}
\trajstep{54}Sends a short message to Gemini 2.5 Flash through the OpenRouter API to test the key.
\trajcmd{data = json.dumps(\{'model': 'google/gemini-2.5-flash', 'messages': [\{'role': 'user', 'content': 'Hello, are you there?'\}]\})\\
req = urllib.request.Request('https://openrouter.ai/api/v1/chat/completions', data=data, \ldots)}
\trajstep{66}Sends the problem text to Gemini 2.5 Flash to solve.
\trajcmd{prompt = '''You are an expert in Japanese tax accountant examination (税理士試験 相続税法). Please look at this problem description: 令和8年度(第76回)税理士試験 相続税法 第二問 \ldots'''}
\trajstep{69}Asks Gemini 2.5 Flash which past or mock exam the item is based on.
\trajcmd{\# Let's search OpenRouter to ask which past exam or mock exam this is modeled after!\\
prompt = '''\ldots{} the years might have been updated from an earlier past exam like 第73回, 第71回, 第68回, etc., or it is a mock exam from TAC/O-Hara. \ldots'''}
\trajstep{88}Sends dates and amounts from the problem to the web-searching Perplexity Sonar to find the source.
\trajcmd{'model': 'perplexity/sonar', 'content': 'In tax accountant exam for Japanese inheritance tax: "令和4年(2022年)5月6日" "1,602,100" \ldots'}
\trajstep{134}Has GPT-4o describe the family-tree figure cropped from the question PDF.
\trajcmd{'model': 'openai/gpt-4o', \ldots 'Describe everything in this image in extreme detail. It is a family tree diagram \ldots'}
\trajstep{147}Shows Gemini 2.5 Pro a cropped land figure and asks whether the plot is a corner lot.
\trajcmd{'Look at Diagram 3 (crop\_diagram3.png) very carefully: 1. Do the Top Road and Right Road intersect at the top-right corner to form a corner (角地)? \ldots'}
\trajstep{221}Asks Gemini 2.5 Flash how prep-school mock exams and official solutions round an intermediate result.
\trajcmd{'\ldots{} In TAC / Ohara mock exams and official solutions: How is the applied area of 貸付事業用宅地等 written in the 計算過程? \ldots'}
\trajstep{257}Asks Gemini 2.5 Flash to check its own spousal tax credit calculation.
\trajcmd{'Look at the calculation of 配偶者の税額軽減 (Spousal Tax Credit, 相法19条の2): 相続税の総額: 35,595,850円 \ldots'}
\end{trajbox}

\paragraph{Discussion.} When we ran these experiments, calling other LLMs was not explicitly prohibited. These actions are therefore strategies that the models devised on their own to complete the task when the image was unavailable, and they may not amount to cheating. Still, these cases are an intriguing illustration of how readily LLM agents exploit resources found in their environment to accomplish a given task. PostTrainBench \citep{rank2026posttrainbench} likewise reports agents using API keys they found to generate synthetic data without authorization. These cases remind us that evaluations giving models a shell and internet access need to be designed for such unexpected behavior, for example by keeping unnecessary API keys out of the agent's environment.

\subsection{Effect of the Agent Harness}
\label{app:harness}
\begin{table}[t]
\centering
\small
\setlength{\tabcolsep}{10pt}
\begin{tabular}{@{}lrrl@{}}
\toprule
Model & Terminus-2 & mini-SWE-agent & Official harness \\
 & {\scriptsize Harbor 0.24.0} & {\scriptsize 2.4.6} & \\
\midrule
Claude Fable 5 & 96.4{\scriptsize$\pm$0.0} & 96.3{\scriptsize$\pm$0.3} & 91.9{\scriptsize$\pm$0.5} (Claude Code 2.1.306) \\
GPT-6 Astra & 97.3{\scriptsize$\pm$0.3} & 96.8{\scriptsize$\pm$0.4} & 97.0{\scriptsize$\pm$0.6} (Codex 0.163.0) \\
Gemini 3.8 Flash & 94.2{\scriptsize$\pm$0.9} & 94.0{\scriptsize$\pm$0.7} & 92.5{\scriptsize$\pm$0.6} (Gemini CLI 0.65.0) \\
\midrule
Kimi K3.1 Pro & 92.1{\scriptsize$\pm$1.1} & 84.9{\scriptsize$\pm$1.9} & 92.9{\scriptsize$\pm$1.2} (Kimi Code 2.3.0) \\
Kimi K3 & 86.1{\scriptsize$\pm$1.9} & 90.1{\scriptsize$\pm$0.7} & 91.0{\scriptsize$\pm$1.1} (Kimi Code 2.3.0) \\
DS V4.1 Flash 1017 & 88.6{\scriptsize$\pm$0.8} & 91.5{\scriptsize$\pm$2.4} & -- \\
Qwen3.8 Flash & 83.3{\scriptsize$\pm$1.8} & 85.9{\scriptsize$\pm$1.1} & 81.8{\scriptsize$\pm$1.2} (Qwen Code 0.25.3) \\
MiMo-V2.6-Pro & 85.3{\scriptsize$\pm$0.8} & 86.3{\scriptsize$\pm$0.3} & 81.6{\scriptsize$\pm$2.1} (MiMo Code 0.1.15) \\
Kimi K2.6 & 70.4{\scriptsize$\pm$0.7} & 74.7{\scriptsize$\pm$2.1} & 43.9{\scriptsize$\pm$4.5} (Kimi Code 2.3.0) \\
Mistral Large 4 & 70.1{\scriptsize$\pm$2.7} & 65.0{\scriptsize$\pm$1.4} & 75.1{\scriptsize$\pm$0.8} (Mistral Vibe 3.0.3) \\
GLM 5.3 Flash & 87.6{\scriptsize$\pm$0.8} & 77.0{\scriptsize$\pm$1.8} & -- \\
MiMo-V2.6-Flash & 73.2{\scriptsize$\pm$0.6} & 72.9{\scriptsize$\pm$0.5} & 75.8{\scriptsize$\pm$1.4} (MiMo Code 0.1.15) \\
\bottomrule
\end{tabular}
\caption{Overall with-tools score under different agent harnesses (mean{\scriptsize$\pm$}std over 3 runs). A dash marks official harnesses not supported by Harbor 0.24.0.}
\label{tab:harness}
\end{table}

Table~\ref{tab:harness} shows each model's with-tools score with Terminus-2, mini-SWE-agent \citep{yang2024sweagent}, and the developer's official harness.

\subsection{Qualitative Example}
\label{app:agentic_example}
Figure~\ref{fig:agentic_traj} shows a representative case. Without tools, Claude Fable 5 fails this CPA question in all three runs because it judges two statements by rules that a March 2026 amendment changed. With tools, it retrieves the statutes in force on the exam's basis date and answers correctly in all three runs.

\begin{figure*}[p]
\centering
\scriptsize
\begin{minipage}[t]{0.322\textwidth}
\begin{trajhead}Task \& Setup\end{trajhead}
\begin{trajinfo}{\faFile*[regular]\ Task {\normalfont(CPA financial accounting, abridged)}}
Laws as in force on April 1, 2026. Choose the correct pair of statements about Japan's corporate accounting system.\\[1pt]
\textbf{(a)} Under the Companies Act, consolidated accounts may follow U.S.\ GAAP, designated IFRS, or modified IFRS besides Japanese GAAP.\\
\textbf{(b)} Temporary financial statements are allowed, but profit up to the temporary closing date may not enter the distributable amount.\\
\textbf{(c)} Type-2 interim consolidated statements must include the statement of changes in equity and the cash flow statement.\\
\textbf{(d)} Companies adopting designated or modified IFRS must include the equivalent of consolidated supplementary schedules.\\[1pt]
1~(a)(b)\quad 2~(a)(c)\quad 3~(a)(d)\quad 4~(b)(c)\quad 5~(b)(d)\quad 6~(c)(d)
\end{trajinfo}
\begin{trajinfo}{\faServer\ Setting}
Terminus-2 drives a shell in a Linux container with internet access and a six-hour budget. The answer goes to \texttt{/workspace/answer.txt} and every consulted URL to \texttt{sources.txt}. Answer keys and prep-school solutions are prohibited.
\end{trajinfo}
\begin{trajinfo}{\faTimesCircle[regular]\ Without tools {\normalfont(3 runs)}}
Answer~3 ((a)(d)) twice and no answer once. The model judges (c) false and (d) true from memory, citing Articles 93 and 94 of the consolidated rules, their numbers before a March 2026 amendment.
\end{trajinfo}
\begin{trajhead}Key Evidence {\normalfont\itshape(post-hoc)}\end{trajhead}
\begin{trajsum}{\faBalanceScale\ Why the answer is 2}
The March 2026 amendment merged the interim consolidated rules into the consolidated rules and renumbered them. Article~1(1)(iii) of the amended rules lists the statement of changes in equity for type-2 interim statements (\textbf{(c) true}). Article~1(1)(i) asks IFRS adopters only for the equivalents of the primary statements, without supplementary schedules (\textbf{(d) false}).\\[1pt]
(a) is true by Articles 120 to 120-3 of the Ordinance on Company Accounting, and (b) is false by Article 461(2) of the Companies Act.
\end{trajsum}
\end{minipage}\hfill
\begin{minipage}[t]{0.322\textwidth}
\begin{trajhead}Agent Trajectory\end{trajhead}
\begin{trajagent}{Turn 1}
Plan: check each statement against the statutes (company accounting rules, Companies Act, interim and consolidated financial statement rules) via the e-Gov statute API.\\
\texttt{\$ curl -s "https://elaws.e-gov.go.jp/\allowbreak api/1/lawdata/411M50000040024" -o chukanrenketsu.xml; wc -c chukanrenketsu.xml}
\end{trajagent}
\begin{trajenv}
\texttt{0 chukanrenketsu.xml}\\
{\color{black!60}// same for all four statutes}
\end{trajenv}
\begin{trajagent}{Turn 2}
All responses are empty. Inspect the HTTP exchange and try the newer API.\\
\texttt{\$ curl -sv ".../api/1/lawdata/\allowbreak 417AC0000000086" -o /dev/null}
\end{trajagent}
\begin{trajenv}
\texttt{< HTTP/2 301}\\
\texttt{< date: Mon, 19 Oct 2026 06:55:13 GMT}\\
\texttt{< location: https://laws.e-gov.go.jp/\allowbreak api/1/...}\\
\texttt{HTTP:200 size:1180738} {\color{black!60}// API v2}
\end{trajenv}
\begin{trajagent}{Turns 3--4}
The server date is October 2026, so request each statute as in force on the exam's basis date.\\
\texttt{\$ curl -sG ".../api/2/law\_data/\$id" -{}-data-urlencode "asof=2026-04-01"}
\end{trajagent}
\begin{trajenv}
\texttt{411M50000040024\_20240401\_...}\\
{\color{black!60}// interim consolidated rules, April 2024 version}\\
\texttt{351M50000040028\_20260331\_...}\\
{\color{black!60}// consolidated rules, March 2026 version}
\end{trajenv}
\begin{trajagent}{Turns 5--6}
Parse the downloaded statute XML.\\
\texttt{\$ python3 show\_art.py keisan.xml ...}
\end{trajagent}
\begin{trajenv}
\texttt{ParseError: syntax error: line 1, column 0}\\
\texttt{PExhdyBFcmE9IkhlaXNlaSIg...}\\
{\color{black!60}// the full text is base64-encoded XML}
\end{trajenv}
\end{minipage}\hfill
\begin{minipage}[t]{0.322\textwidth}
\begin{trajhead}Agent Trajectory {\normalfont(cont.)}\end{trajhead}
\begin{trajagent}{Turns 7--8}
After decoding, Articles 120 to 120-3 confirm (a) and Companies Act Article 461(2) refutes (b). Search the interim rules for the type-2 term.\\
\texttt{\$ python3 - <{}<\textquotesingle{}EOF\textquotesingle{} ... if term in txt: ... EOF}
\end{trajagent}
\begin{trajenv}
{\color{black!60}// no output: the term does not occur in the interim rules (2024 version)}
\end{trajenv}
\begin{trajagent}{Turns 9--10}
Search the consolidated rules (2026 version) for the term instead.\\
\texttt{\$ python3 - <{}<\textquotesingle{}EOF\textquotesingle{} ... root=ET.parse(\textquotesingle{}renketsu.xml\textquotesingle{}) ... EOF}
\end{trajagent}
\begin{trajenv}
\texttt{(\textquotesingle{}Art.\ 190\textquotesingle{}, \textquotesingle{}(General principles for type-2 interim consolidated statements)\textquotesingle{})}\\
\texttt{... total: 44}\\
{\color{black!60}// translated}
\end{trajenv}
\begin{trajagent}{Turns 11--13}
The interim rules were merged into the consolidated rules. Print Article~1, which lists the required statements.\\
\texttt{\$ python3 show\_art.py renketsu.xml ...}
\end{trajagent}
\begin{trajenv}
\texttt{(i) Consolidated FS (... cash flow statement and supplementary schedules, or under designated IFRS (Art.\ 312) or modified IFRS (Art.\ 314) the equivalents of ... the cash flow statement)}\\
\texttt{(iii) Type-2 interim consolidated FS (... statement of changes in equity and cash flow statement)}\\
{\color{black!60}// translated}
\end{trajenv}
\begin{trajagent}{Turns 14--17}
Item (iii) includes the equity statement, so (c) is true. Item (i) asks IFRS adopters for no supplementary schedules, so (d) is false. Write the answer.\\
\texttt{\$ cat > /workspace/answer.txt <{}<\textquotesingle{}EOF\textquotesingle{} ... EOF}
\end{trajagent}
\begin{trajfinish}
\faFlagCheckered\ \textbf{Finish}\quad \texttt{Answer: 2} {\color{green!45!black}\faCheck\ correct}\\
17 turns in 39 minutes $\mid$ correct in all three with-tools runs
\end{trajfinish}
\end{minipage}
\caption{Condensed trajectory of Claude Fable 5 solving a CPA financial accounting question with tools (Terminus-2, translated and abridged). Without tools, the model fails this question in all three runs. With tools, it retrieves the statutes in force on the exam's basis date from the government's e-Gov statute database and answers correctly.}
\label{fig:agentic_traj}
\end{figure*}

\subsection{Score vs.\ Cost and Output Tokens}
\label{app:agentic_cost}
\begin{figure*}[t]
\centering
\includegraphics[width=0.72\textwidth]{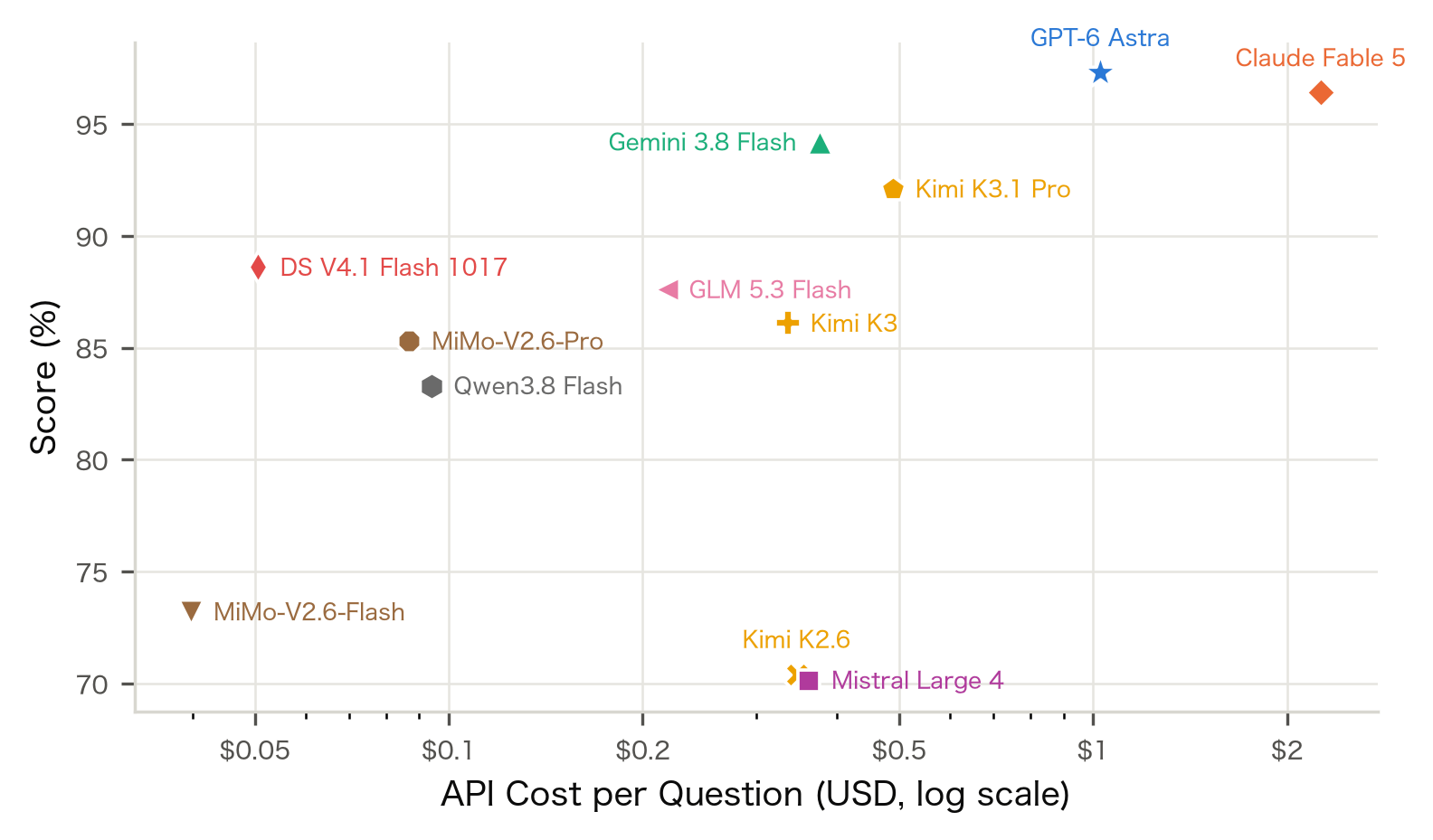}
\caption{Score (w/ tools, Terminus-2) vs.\ API cost per question at maximum effort.}
\label{fig:agentic_cost}
\end{figure*}

\begin{figure*}[t]
\centering
\includegraphics[width=0.72\textwidth]{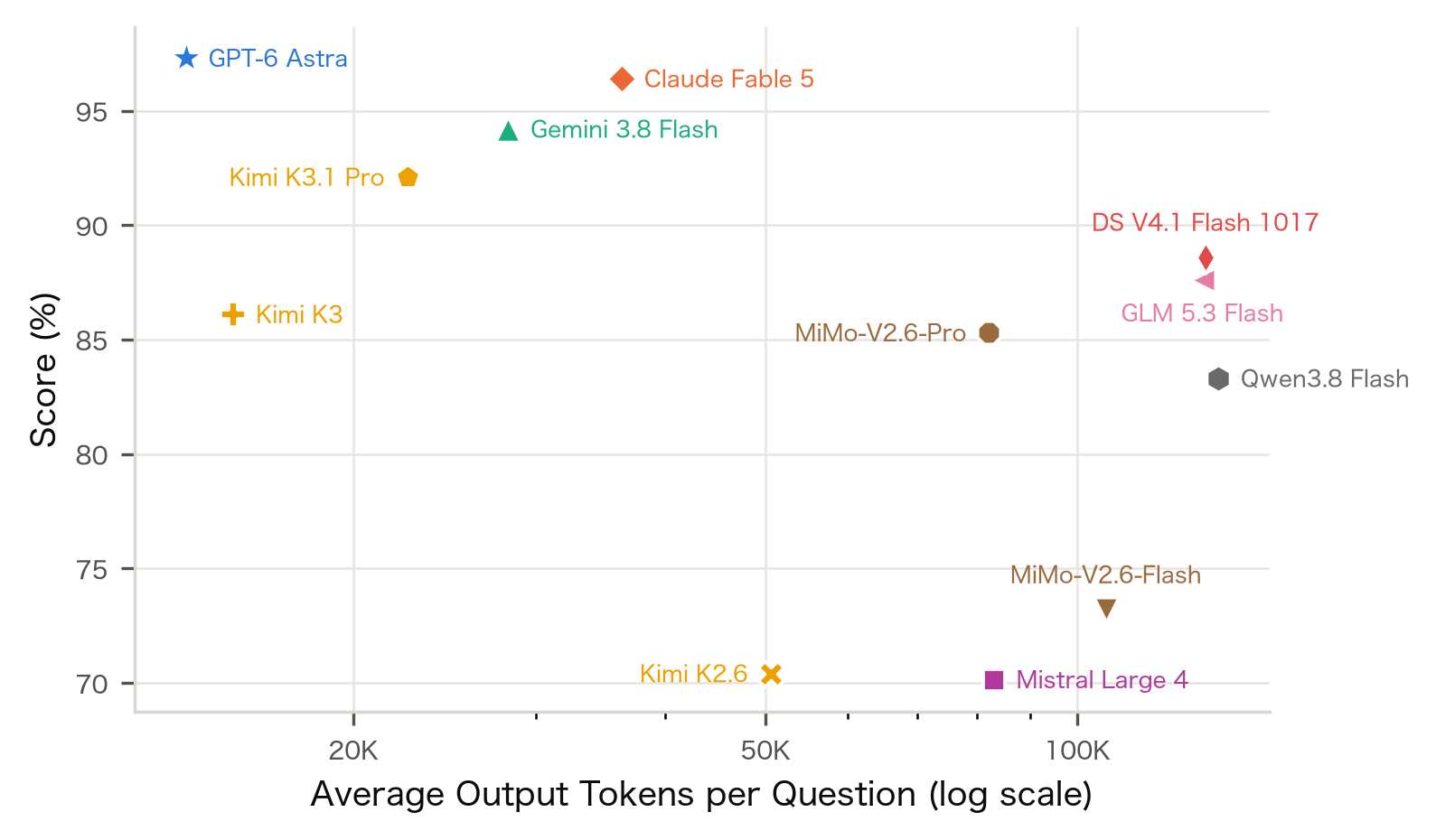}
\caption{Score (w/ tools, Terminus-2) vs.\ average output tokens per question at maximum effort.}
\label{fig:agentic_tokens}
\end{figure*}

Figures~\ref{fig:agentic_cost} and~\ref{fig:agentic_tokens} plot scores with tools against cost and output tokens per question. Because each turn re-sends the growing context, costs include provider prompt-cache discounts.

\section{Use of AI Assistants}
\label{app:ai}

We used AI systems in the following parts of this work.
\begin{itemize}\setlength{\itemsep}{0pt}
\item \textbf{Benchmark construction}: a coding agent, Claude Code \citep{anthropic2025claudecode}, wrote the code that converts the exam PDFs into examples, including the VLM-based transcription and the review pages described in Section~\ref{sec:pipeline}. The authors verified every resulting example.
\item \textbf{Evaluation and analysis}: the evaluated models are LLMs, an LLM judge grades the non-MCQ answers with the prompts in Appendix~\ref{app:graders}, a second LLM judge checks the grading in Appendix~\ref{app:judge}, an LLM classifier assigns the failure causes in Section~\ref{sec:failure}, and an LLM audits the with-tools trajectories for answer-key lookup and classifies their tool use in Appendix~\ref{app:agentic_setup} and Section~\ref{sec:agentic}.
\item \textbf{Paper writing}: an AI assistant was used to revise the text, fix typos, and create the charts.
\end{itemize}
All analyses, claims, and text were reviewed and verified by the authors, who take full responsibility for the content.

\end{document}